\documentclass[11pt]{article}

\usepackage[preprint]{acl}

\usepackage{times}
\usepackage{latexsym}

\usepackage[T1]{fontenc}

\usepackage[utf8]{inputenc}

\usepackage{microtype}

\usepackage{inconsolata}

\usepackage{graphicx}
\usepackage{float}
\usepackage{fvextra}
\usepackage{amsmath,amsfonts,bm}

\def\eqref#1{equation~\ref{#1}}

\def\1{\bm{1}}

\DeclareMathAlphabet{\mathsfit}{\encodingdefault}{\sfdefault}{m}{sl}
\SetMathAlphabet{\mathsfit}{bold}{\encodingdefault}{\sfdefault}{bx}{n}

\usepackage{kotex}

\usepackage{ulem}   
\newcommand\highlight[1][yellow]{%
  \bgroup 
  \markoverwith{\textcolor{#1}{\vrule width.15em height.85em depth.2em}}%
  \ULon 
}
\definecolor{colordhq}{HTML}{8E44AD}
\newcommand{\dhq}[1]{\highlight[colordhq!40]{#1}}
\definecolor{colordqq}{HTML}{3B9AB2}
\newcommand{\dqq}[1]{\highlight[colordqq!40]{#1}}
\definecolor{colorde}{HTML}{E56B55}
\newcommand{\de}[1]{\highlight[colorde!40]{#1}}
\definecolor{colorddq}{HTML}{3F784C}
\newcommand{\ddq}[1]{\highlight[colorddq!40]{#1}}

\usepackage{hyperref}
\usepackage{url}

\usepackage{multirow}
\usepackage{booktabs}

\usepackage{tcolorbox}
\usepackage{listings}
\usepackage{enumitem}

\definecolor{colorpatient}{HTML}{2E5C8A}
\definecolor{colordoctor}{HTML}{A0522D}
\definecolor{coloreval}{HTML}{555555}
\tcbset{
  patientprompt/.style={colframe=colorpatient, colback=colorpatient!5},
  doctorprompt/.style={colframe=colordoctor, colback=colordoctor!5},
  evalprompt/.style={colframe=coloreval, colback=coloreval!5},
}

\newcommand{\ours}{\textsc{DIPSY}}

\title{
\ours{}: Evaluating Diagnostic Information Acquisition in LLM-Based Psychiatric Interviews}

\newcommand\CoauthorMark{\footnotemark[\arabic{footnote}]}
\newcommand\CorrMark{\footnotemark[\arabic{footnote}]}
\author{
 \textbf{Jisu Shin\textsuperscript{1}\thanks{Equal contribution}} \hspace{6mm}
 \textbf{Migyeong Kang\textsuperscript{2}\CoauthorMark} \hspace{6mm}
 \textbf{Jihyun Kim\textsuperscript{2}} \hspace{6mm}
 \textbf{Hangyeol Yoo\textsuperscript{1}} \hspace{6mm}
\\
 \textbf{KyungTae Lim\textsuperscript{1}} \hspace{6mm}
 \textbf{Jihyun An\textsuperscript{3}} \hspace{6mm}
 \textbf{Alice Oh\textsuperscript{1}} \hspace{6mm}
 \textbf{Jinyoung Han\textsuperscript{2}}\thanks{Corresponance} \hspace{6mm}
 \textbf{Hoyun Song\textsuperscript{4}\CorrMark}
\\
 \textsuperscript{1}KAIST \hspace{2mm}
 \textsuperscript{2}Sungkyunkwan University \hspace{2mm}
 \textsuperscript{3}Samsung Medical Center \hspace{2mm}
 \textsuperscript{4}Dankook University
\\
 \small{
   \texttt{jisu.shin@kaist.ac.kr \hspace{2mm} gy77@g.skku.edu \hspace{2mm} jinyounghan@skku.edu \hspace{2mm} hoyun.song@dankook.ac.kr}
 }
}

\begin{document}
\maketitle

\begin{abstract}
Large language models (LLMs) are increasingly evaluated in multi-turn mental health assessment, where diagnosis depends not only on interpreting the information already available, but also on identifying what remains unknown and asking appropriate questions to obtain it. We refer to this capability as \textbf{diagnostic information acquisition} (DIA). However, existing mental health benchmarks typically present models with fixed diagnostic information and focus on final diagnostic outcomes, offering limited insight into how models acquire evidence through interaction. We introduce \ours{}, a controlled framework for evaluating DIA in multi-turn psychiatric interviews. \ours{} uses a profile-grounded patient simulator that selectively discloses diagnostic evidence in response to the doctor model's questions, allowing the acquisition process to be tracked throughout the dialogue. It evaluates the doctor along four dimensions: hypothesis quality, question quality, efficiency, and decision quality. Across multiple doctor LLMs and evaluation settings, we find that models with similar final accuracy exhibit substantial differences in hypothesis coverage, question effectiveness, interview efficiency, and evidence sufficiency.\footnote{ All resources will be publicly released: \url{https://anonymous.4open.science/r/DIPSY-80B3}}
\end{abstract}

\section{Introduction}
\begin{figure}[t]
    \centering
    \includegraphics[width=\columnwidth]{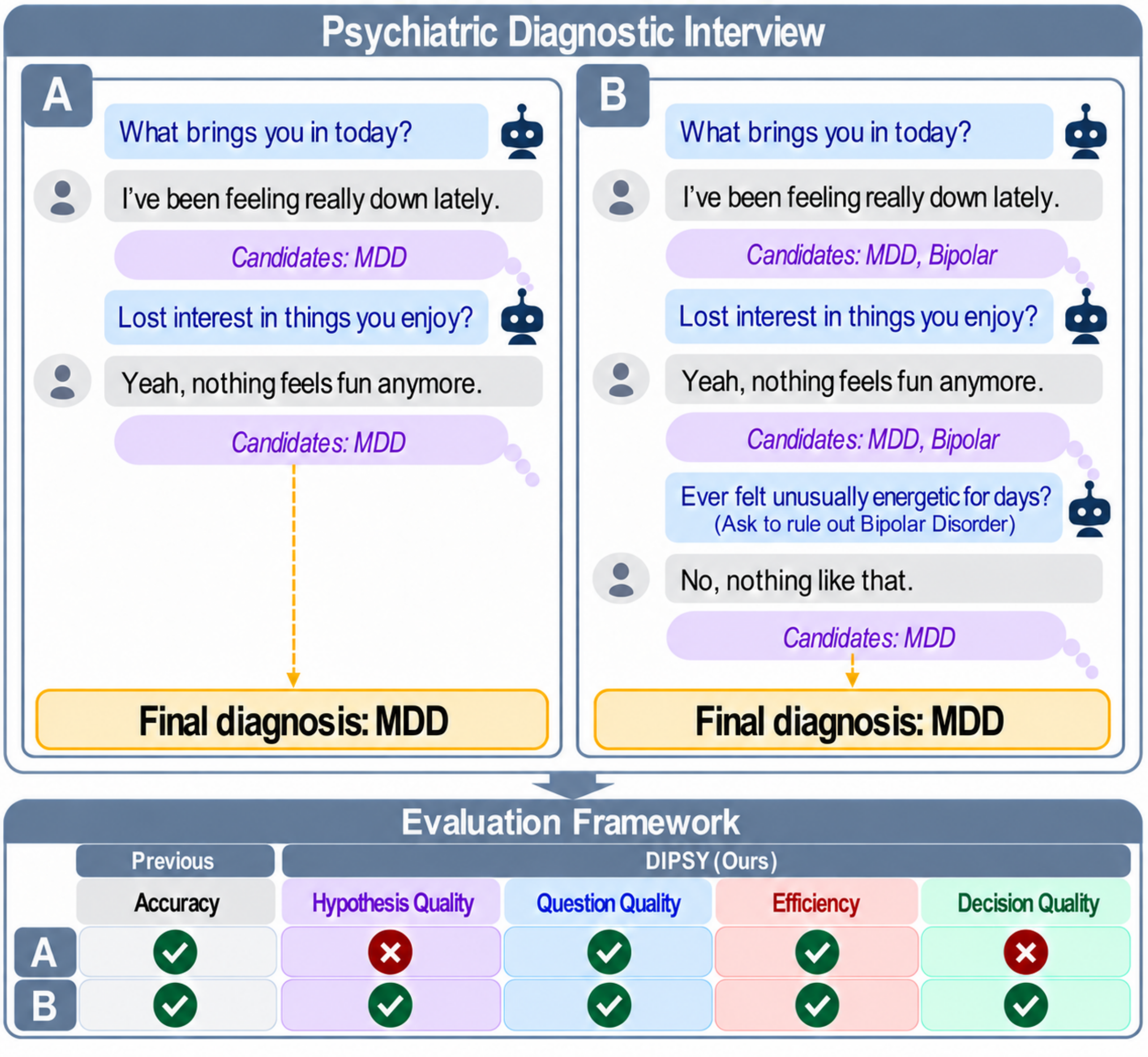}
    \caption{\label{fig:concept}Example psychiatric interviews with different diagnostic processes and evaluation outcomes.}
    \vspace{-0.25in}
\end{figure}

Large language models (LLMs) are increasingly assessed for capabilities relevant to psychiatric diagnosis~\cite{xu2026lingxidiagbench, fouda2026psychiatrybench, song2026mentalbench, 
liu2025psychbench, sarma2025integrating}. In psychiatric interviews, diagnosis often unfolds through conversation rather than through biomarkers or written records. Clinically relevant information may be disclosed only partially, requiring clinicians to elicit needed evidence through targeted questions~\cite{savander2021patients, savander2024take, mcallister2016clinical}. Accordingly, LLMs conducting such interviews need to identify missing information and ask follow-up questions to obtain it, rather than relying solely on information already provided. In this work, we focus on evaluating this capability, which we term \textbf{diagnostic information acquisition (DIA)}.

In clinical practice, DIA proceeds through interconnected stages. Clinicians form plausible diagnostic hypotheses, refine them through further questioning, gather sufficient evidence efficiently, and ultimately reach a diagnosis~\cite{lazare1976psychiatric, nurcombe1982psychiatrists, faust1989improving, holmes2011revisiting}. 
Evaluating only the final diagnosis therefore provides an incomplete view of this process.
As illustrated in \autoref{fig:concept}, both interviews reach the same correct diagnosis, but Case B shows a clearer intermediate process by considering alternatives and acquiring additional evidence. By contrast, Case A offers less insight into how the decision is reached, making it difficult to explain which evidence supports the final diagnosis. Therefore, process-level evaluation is needed to characterize the strengths and weaknesses of the diagnostic trajectory before the final decision.

However, existing evaluations are not well suited for measuring DIA due to two key challenges. First, evaluating DIA requires controlling how diagnostic evidence is disclosed in response to the model’s questions. Without such control, models may receive varying amounts of diagnostic evidence regardless of question quality, making it difficult to assess how effectively their questions elicit useful information. Second, intermediate diagnostic states must be evaluated against the information available at each point in the interview. Unlike the final diagnosis, which is evaluated using the complete dialogue, intermediate hypotheses and follow-up questions cannot be evaluated against evidence that has not yet been revealed.

To address these challenges, we introduce \ours{}, a framework for \textbf{D}iagnostic \textbf{I}nterview \textbf{P}rocess Evaluation in \textbf{Psy}chiatry. 
\ours{} uses a patient simulator with a hidden clinical profile and selectively discloses diagnostic evidence in response to the doctor model’s questions. This controlled interaction allows us to track the evidence elicited at each turn and link it to the corresponding question. We further propose metrics that evaluate four stages of the DIA process: diagnostic hypothesis formation, question selection, evidence acquisition efficiency, and final diagnostic decision-making. This process-level evaluation enables fine-grained identification of failure modes along the diagnostic trajectory.

Our experimental results show that models with similar final diagnostic accuracy can exhibit substantially different behaviors throughout the diagnostic process. Using our framework, we identify differences in how models maintain diagnostic hypotheses, select follow-up questions, narrow candidate sets, and determine when to stop the interview. These findings show that process-level evaluation reveals diagnostic behaviors and failure modes that final accuracy alone cannot capture. 
Our contributions are summarized as follows:

\begin{itemize}[leftmargin=*,topsep=-2px,partopsep=0px]
    \item We introduce a controlled framework for evaluating DIA in multi-turn psychiatric interviews, where diagnostic evidence is selectively disclosed in response to model questions, enabling explicit tracking of the acquisition process.

    \item We propose process-level metrics that evaluate DIA throughout the diagnostic trajectory, enabling fine-grained identification of failures in hypothesis formation, question selection, evidence acquisition, and final decision-making.

    \item Our experiments reveal diverse diagnostic behaviors across models and pinpoint where each is vulnerable in the diagnostic process, which final accuracy alone cannot identify.
\end{itemize}

\section{Related Work}
\subsection{Interactive Psychiatric Diagnosis}

Existing mental health benchmarks have largely relied on static evaluation settings, such as disorder classification~\cite{wang2024explainable, kang2024cure, kabir2023deptweet, coppersmith2015clpsych}, clinical vignette interpretation~\cite{fouda2026psychiatrybench, sarma2025integrating}, and question answering~\cite{racha2025mhqa, alhuzali2024mentalqa}.
Because static benchmarks typically provide diagnostically relevant information in advance, they are less suited to evaluating diagnostic processes that unfold through interaction.

More recent work has extended psychiatric diagnosis to interactive dialogue using LLM-based patient simulators~\cite{xu2026lingxidiagbench, wu2026wisemind,yin2025mdd, ozgun2025trustworthy}. These simulators are commonly built from patient records, symptom profiles, or personas to generate realistic and profile-consistent responses~\cite{kyung2026patientsim, lee2025psyche, wang2024patient}. However, they are not typically designed to control information disclosure for DIA evaluation, and diagnostic information may therefore be revealed beyond what is directly elicited by the doctor model’s questions. Our framework instead uses patient simulation as a controlled evaluation environment, where the full clinical profile remains hidden, and information is selectively disclosed in response to the doctor model’s questions.

\begin{figure*}[t]
    \centering
    \includegraphics[width=1\linewidth]{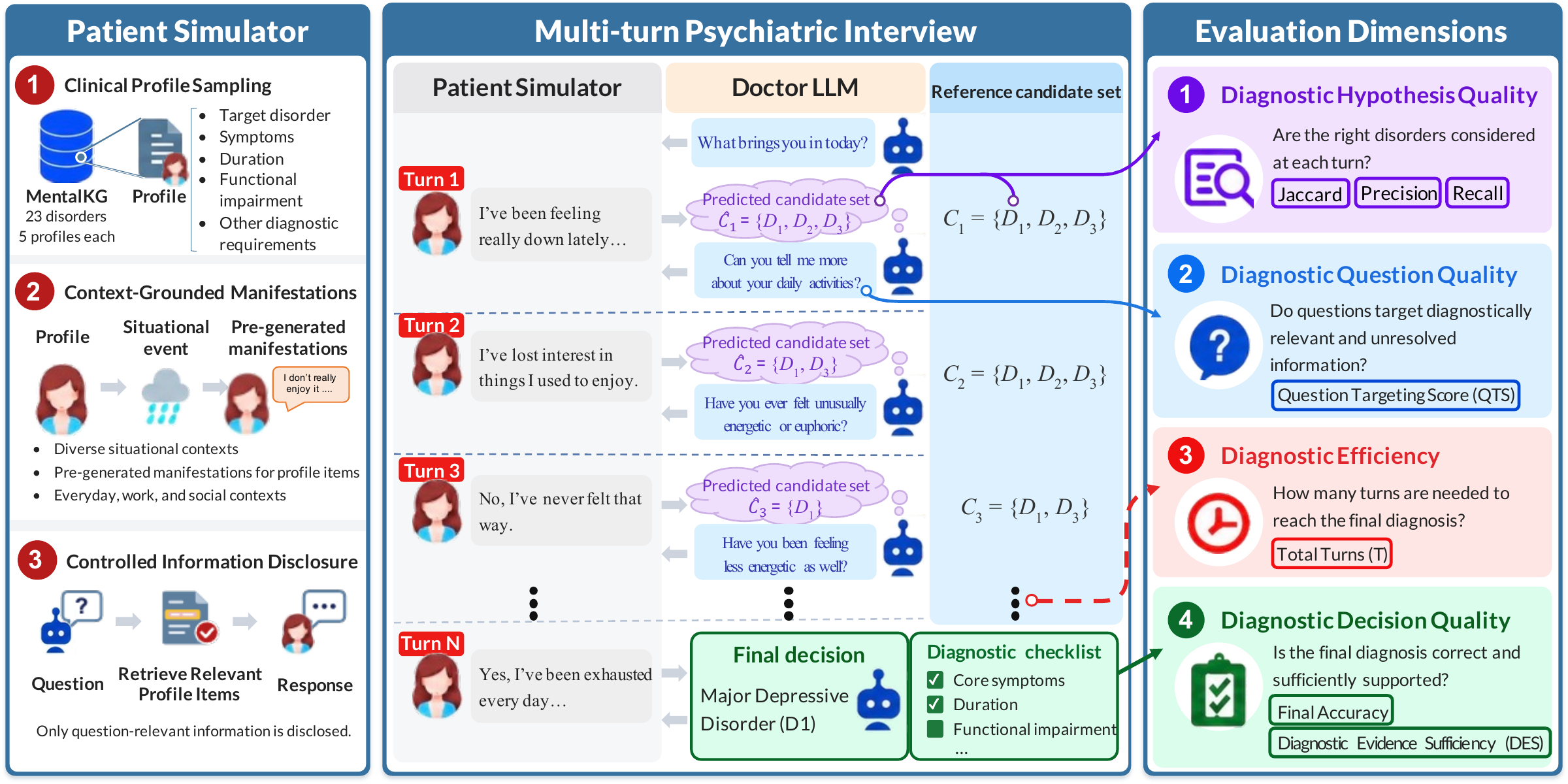}
    \caption{\label{fig:framework}Overview of \ours{}. A doctor LLM interacts with a profile-grounded patient simulator, while the process is evaluated across four dimensions: hypothesis quality, question quality, efficiency, and decision quality.}
    \vspace{-0.2in}
\end{figure*}

\subsection{Process-Level Evaluation of Diagnostic Dialogue}

Existing evaluations of psychiatric diagnosis have primarily focused on final outcomes, such as diagnostic accuracy~\cite{lan2024depression, jin2026psyeval} or disorder identification~\cite{xu2025identifying}. Some studies have also evaluated the quality or sufficiency of information gathering through post-hoc assessment of complete transcripts or with LLM-based judges~\cite{xu2026lingxidiagbench, wu2026wisemind, chandra2026reasoning}. While these approaches can assess the overall quality of a diagnostic interview, they provide limited insight into how well the diagnostic process is grounded before the final decision.

Prior work has also examined turn-level evaluation of diagnostic dialogue~\cite{xu2026lingxidiagbench}. However, these studies typically assess the appropriateness of individual responses or questions at each turn, rather than whether the diagnostic process remains coherent across turns. Our work instead evaluates the DIA process relative to the evidence accumulated at each turn and proposes metrics for four aspects of the diagnostic trajectory: diagnostic hypothesis quality, diagnostic question quality, diagnostic efficiency, and diagnostic decision quality. This enables fine-grained analysis of where failures occur during diagnostic information acquisition.

\section{\ours{}}

\subsection{Diagnostic Interview Protocol}

We introduce \ours{}, a framework for evaluating how doctor LLMs acquire diagnostic information during multi-turn psychiatric interviews. Each interview is conducted between a doctor LLM as the evaluation target and a patient simulator provided by our framework. The framework controls information disclosure through the patient simulator and evaluates the doctor's diagnostic process across multiple stages.

The interview begins with the doctor's opening question, \textit{``What brings you in today?''}, to which the patient responds with their chief complaint. As illustrated in Figure~\ref{fig:framework}, at each subsequent turn $t$, the doctor reports its current diagnostic hypotheses as a candidate set $\hat{C}_t$ and either asks a follow-up question $q_t$ or decides to make a final diagnosis. The patient then responds to $q_t$, and the doctor updates its hypotheses based on the accumulated dialogue. When the doctor decides to stop, or when the maximum turn limit is reached, it submits a final diagnosis and an evidence checklist summarizing the information supporting its decision. Throughout the interview, the doctor has access only to the dialogue history, its own previous candidate sets, and its own diagnostic knowledge; the patient's hidden clinical profile and target disorder are not available to the doctor. Implementation details are provided in Appendix~\ref{app:framework_doctor}.

The doctor's outputs are evaluated along four dimensions. The candidate set $\hat{C}_t$ is used to assess \textsc{\dhq{Diagnostic Hypothesis Quality}}, the follow-up question $q_t$ to assess \textsc{\dqq{Diagnostic Question Quality}}, the interview length to assess \textsc{\de{Diagnostic Efficiency}}, and the final diagnosis and evidence checklist to assess \textsc{\ddq{Diagnostic Decision Quality}}. An LLM-based judge extracts structured diagnostic information from the dialogue for these evaluations.

\subsection{Controlled Patient Simulation}
\label{sec:patient_simulator}

\paragraph{Clinical Profile Sampling.}
The patient simulator is grounded in a clinical profile sampled from MentalKG~\citep{song2026mentalbench}, a DSM-grounded knowledge graph. The profiles cover 23 psychiatric disorders, with five distinct profiles per disorder, for a total of 115 profiles. Each profile specifies the target disorder, its symptoms, duration, and other disorder-specific diagnostic requirements. This profile serves as the patient's hidden clinical state, from which relevant information is gradually disclosed during the interview.

\paragraph{Context-Grounded Symptom Manifestation.}
To express diagnostic information without directly exposing clinical terminology, we construct patient-specific contexts and prepare a manifestation for each profile item in advance. Each profile is instantiated as three patients using distinct personas sampled from Nemotron-Personas-USA~\citep{nvidia/Nemotron-Personas-USA}, yielding 345 patients in total. Each persona is assigned a stressful life event from the Holmes--Rahe Social Readjustment Rating Scale~\citep{holmes1967social}, which is combined with the clinical profile to generate an onset narrative.

Based on this context, we generate a first-person manifestation for each profile item that describes the patient's everyday experience without explicitly naming the underlying clinical concept. This applies to symptoms as well as other diagnostic requirements, including duration and functional impairment. One symptom is selected as the chief complaint and used as the patient's first-turn response, providing a limited initial presentation that requires further questioning. Further implementation details are provided in Appendix~\ref{app:patient_implementation}.

\paragraph{Controlled Information Disclosure.}
To control information disclosure, we generate each patient response in two stages. In the \textit{alignment} stage, the simulator identifies which clinical variable from the hidden profile is targeted by the doctor's question $q_t$; if no variable can be identified, it returns \textit{none}. In the \textit{generation} stage, only the aligned variable and its pre-generated manifestation are provided to the response generator, which produces a reply consistent with the dialogue context. If \textit{none} is returned, the patient is instructed to express uncertainty or report its absence, without disclosing any profile information. 
Because no other clinical content is provided, newly revealed evidence can be linked to the question that elicited it.


\begin{table}[t]
\centering
\resizebox{\columnwidth}{!}{%
\begin{tabular}{lcccc}
\toprule
\textbf{Dimension} & \textbf{Expert} & \textbf{LLM} & \textbf{Exact} & \textbf{Gwet's AC$_2$} \\
\midrule
Profile Realism      & 4.85 & 4.98 & 87\% & 0.99 \\
Conversation Realism & 4.65 & 4.85 & 76\% & 0.97 \\
Profile Fidelity     & 4.72 & 4.83 & 74\% & 0.97 \\
\bottomrule
\end{tabular}
}
\caption{\label{tab:expert_validation}Expert and LLM evaluation of the patient simulator. Expert denotes the mean of majority-voted ratings from three experts, and LLM denotes the mean rating from Claude Sonnet 5, both on a 1--5 scale. We additionally report the exact-match rate and Gwet's AC$_2$ with quadratic weights between the two.}
\vspace{-0.2in}
\end{table}

\paragraph{Quality Evaluation.}
We conduct an expert evaluation to assess whether the simulated patient presentation is consistent with clinical practice. Three clinical psychologists evaluate 46 conversations, two from each of the 23 disorders, for profile realism, conversation realism, and profile fidelity. As shown in Table~\ref{tab:expert_validation}, all three dimensions receive high ratings, supporting the clinical plausibility of gradual, profile-consistent information disclosure. To extend the evaluation at scale, we use Claude Sonnet 5 as an automated evaluator under the same criteria, which shows high agreement with the expert ratings. Across the full set, the simulator receives high mean ratings of 4.98, 4.78, and 4.81 on the three dimensions, respectively, indicating consistently high quality beyond the expert sample. Further details are provided in Appendix~\ref{app:patient_quality}.

\subsection{Process Evaluation}
\label{sec:process_evaluation}
As diagnostic evidence accumulates, the doctor's hypotheses and follow-up questions are evaluated based on the information available at each turn. An LLM-based evaluation judge extracts the targeted diagnostic variables and revealed evidence from each turn, which are then used with MentalKG to compute the process-level metrics. Diagnostic efficiency and final decision quality are assessed over the complete interview.

\paragraph{Turn-Level Reference Candidate Set.}
At each turn, we construct a reference candidate set based on the diagnostic evidence revealed by the patient up to that point. Let $E_t$ denote the accumulated evidence up to turn $t$. MentalKG links each disorder to its symptoms and other diagnostic requirements, including duration, functional impairment, and disorder-specific criteria. Using these relations and $E_t$, we construct $C_t$ as the set of disorders that remain compatible with the patient's revealed evidence. As new information is disclosed, $C_t$ is updated accordingly. This reference set is then compared with the doctor's predicted candidate set $\hat{C}_t$. Construction rules are provided in Appendix~\ref{app:reference_candidates}.

\paragraph{\textsc{\dhq{Diagnostic Hypothesis Quality.}}}
This dimension evaluates whether the doctor maintains an appropriate set of diagnostic hypotheses. At each turn $t$, we compare the doctor's predicted candidate set $\hat{C}_t$ with the reference set $C_t$. We report \textbf{Jaccard} as the primary metric
to capture both unsupported inclusions and premature exclusions.
We additionally report \textbf{Precision} and \textbf{Recall} to distinguish these two error types. For each metric, we first average the turn-level scores within each interview and then average these interview-level scores across the evaluated interviews.

\paragraph{\textsc{\dqq{Diagnostic Question Quality.}}}
This dimension evaluates whether the doctor's follow-up question appropriately targets diagnostically useful information given its current hypotheses $\hat{C}_t$. For each question, we identify the symptoms it targets and judge each target as useful if it helps distinguish among the current candidate disorders or corresponds to a mandatory diagnostic requirement of one of them, based on the differential-diagnosis relations and disorder-specific requirements in MentalKG. The \textbf{Question Targeting Score (QTS)} evaluates each targeted variable separately with a binary score and normalizes the scores so that each turn contributes at most one point. Questions that revisit information already confirmed or ruled out are not rewarded. Formal definitions are provided in Appendix~\ref{app:question_targeting_score}.

\paragraph{\textsc{\de{Diagnostic Efficiency.}}}
This dimension evaluates how efficiently the doctor acquires diagnostic information before reaching a final diagnosis. We report \textbf{Total Turns ($T$)}, defined as the number of patient-response turns before the doctor submits its final diagnosis. Lower values indicate shorter interviews; when interpreted together with the other process metrics, they help identify premature termination or unnecessarily prolonged questioning.

\paragraph{\textsc{\ddq{Diagnostic Decision Quality.}}}
This dimension evaluates the correctness of the final diagnosis and the sufficiency of its supporting evidence. We report \textbf{Final Accuracy} based on whether the diagnosis matches the target disorder, and \textbf{Diagnostic Evidence Sufficiency (DES)}, which measures the proportion of required evidence for the predicted disorder that has been collected during the interview. DES combines symptom-criteria coverage from symptoms confirmed or denied during the dialogue with fulfillment of non-symptom requirements, such as duration, functional impairment, and disorder-specific criteria, using the doctor's evidence checklist and MentalKG. Because DES is computed for the predicted disorder regardless of correctness, it measures how well the doctor's decision is supported by the acquired evidence. Implementation details are provided in Appendix~\ref{app:decision_quality}.

\begin{table}[t]
    \centering
    \small
    \resizebox{\columnwidth}{!}{
    \begin{tabular}{ll|ll|ll}
    \toprule
        &  & \multicolumn{4}{c}{\textbf{Expert / Automated Ratings}} \\
    \cmidrule{3-6}
        & \textbf{Metrics} &  \multicolumn{2}{c|}{Hypothesis Q.} &  \multicolumn{2}{c}{Question Q.} \\
    \midrule
       \multirow{2}{*}{\textbf{Ours}} & Jaccard & 0.43*** & / 0.55*** & 0.07 &/ 0.04 \\
    \cmidrule{2-6}
        & QTS & 0.06 & / 0.01 & 0.46*** & / 0.52*** \\
    \bottomrule
    \multicolumn{6}{l}{*: p<0.05, **: p<0.01, ***: p<0.001}
    \end{tabular}
    }
    \caption{Pearson correlations ($r$) of our metrics with human-expert and automated (LLM-as-a-judge) ratings of hypothesis and question quality.}
    \label{tab:expert_validation_judge}
    \vspace{-0.1in}
\end{table}
\paragraph{Expert Validation of Process Metrics.}
Two psychiatrists evaluate 397 diagnostic hypothesis sets and 351 follow-up questions from 46 interview episodes on 5-point Likert scales. We compute Pearson correlations between our metrics and the mean ratings of the two experts.
Inter-rater agreement, measured using the intraclass correlation coefficient (\citealp{koo2016guideline}; ICC(2,1)), is 0.63 for hypothesis quality and 0.77 for question quality.
As shown in \autoref{tab:expert_validation_judge}, Jaccard and QTS correlate positively with expert-rated hypothesis and question quality, respectively ($r=0.43$ and $0.46$; both $p<0.001$). Ratings from Claude Sonnet 5 show consistent associations ($r=0.55$ and $0.52$, respectively; both $p<0.001$). These results support the alignment of our metrics with expert judgments of their corresponding dimensions. Detailed evaluation procedures are provided in Appendix~\ref{app:metric_validation}.
\section{Experiments}
\label{sec: Experiments}

\paragraph{Experimental Setup.}
We use two LLMs, GPT 5.6 Terra~\cite{openai2026gpt56} and Gemini 3.1 Pro~\cite{google2026gemini31pro}, to establish two evaluation settings. In each setting, the same model serves as both the patient simulator and evaluation judge, while the doctor model varies. We evaluate five doctor models: three closed-source models (GPT 5.4~\cite{openai2026gpt54}, Gemini 3.8 Flash~\cite{google2026gemini38flash}, and Claude Sonnet 5~\cite{anthropic2026claude}) and two open-source models (Llama 3.3 70B Instruct~\cite{grattafiori2024llama} and Qwen3 235B~\cite{yang2025qwen3}). Each interview is limited to 20 turns.

\begin{table*}[t]
\centering
\small
\renewcommand{\arraystretch}{1.00}
\resizebox{0.9\textwidth}{!}{%
\begin{tabular}{ll|rrr|r|r|rr}
\toprule
 &
   &
  \multicolumn{3}{c|}{\dhq{Hypothesis Q.}} &
  \multicolumn{1}{c|}{\dqq{Question Q.}} &
  \multicolumn{1}{c|}{\de{Efficiency}} &
  \multicolumn{2}{c}{\ddq{Decision Q.}} \\
Patient \& Judge &
  Doctor &
  \multicolumn{1}{c}{Jacc.} &
  \multicolumn{1}{c}{Prec.} &
  \multicolumn{1}{c|}{Rec.} &
  \multicolumn{1}{c|}{QTS} &
  \multicolumn{1}{c|}{$T$ ($\downarrow$)} &
  \multicolumn{1}{c}{DES} &
  \multicolumn{1}{c}{Acc.} \\
\midrule
\multirow{5}{*}{GPT 5.6 Terra} &  GPT 5.4                & 0.43 & 0.57 & \textbf{0.65} & 0.43 & 15.06 & \textbf{0.73}                 & 73.91 \\
                                        & Gemini 3.8 Flash       & 0.46 & 0.66 & 0.62 & \textbf{0.55} & \textbf{7.30}  & 0.67                 & 71.14 \\
                                        & Claude Sonnet 5        & \textbf{0.48} & \textbf{0.72} & 0.59 & 0.47 & 8.41  & 0.69                 & \textbf{75.94} \\
                                        & Llama 3.3 70B & 0.40 & 0.61 & 0.48 & 0.24 & 19.37 & 0.63                 & 41.45 \\
                                        & Qwen3 235B             & 0.33 & 0.52 & 0.46 & 0.32 & 11.50 & 0.58                 & 46.09 \\
\midrule
  \multirow{5}{*}{Gemini 3.1 Pro} & GPT 5.4                & 0.42 & 0.57 & 0.62 & 0.44 & 10.57 & 0.70                 & 69.57 \\
                                        & Gemini 3.8 Flash       & 0.48 & 0.65 & \textbf{0.64} & \textbf{0.50} & \textbf{6.19}  & 0.69                 & 73.62 \\
                                        & Claude Sonnet 5        & \textbf{0.49} & \textbf{0.69} & 0.60 & 0.48 & 6.93  & \textbf{0.72}                 & \textbf{77.39} \\
                                        & Llama 3.3 70B & 0.40 & 0.55 & 0.47 & 0.20 & 18.89 & 0.65                 & 45.80 \\
                                        & Qwen3 235B             & 0.34 & 0.47 & 0.48 & 0.30 & 10.38 & 0.60                 & 50.87 \\
\bottomrule
\end{tabular}%
}
\caption{\label{tab:main_table}Overall performance of doctor models under two evaluation settings. Metrics cover four evaluation dimensions: hypothesis quality, question quality, efficiency, and decision quality. Lower total turns indicate better efficiency, and the best score for each metric within each setting is shown in \textbf{bold}.}
\vspace{-0.1in}
\end{table*}

\subsection{Overall Results}
Our results show that models with similar diagnostic accuracy can exhibit substantially different diagnostic behaviors throughout the interview, highlighting the limitations of outcome-only evaluation.

\begin{figure}[t]
    \centering
    \includegraphics[width=\linewidth]{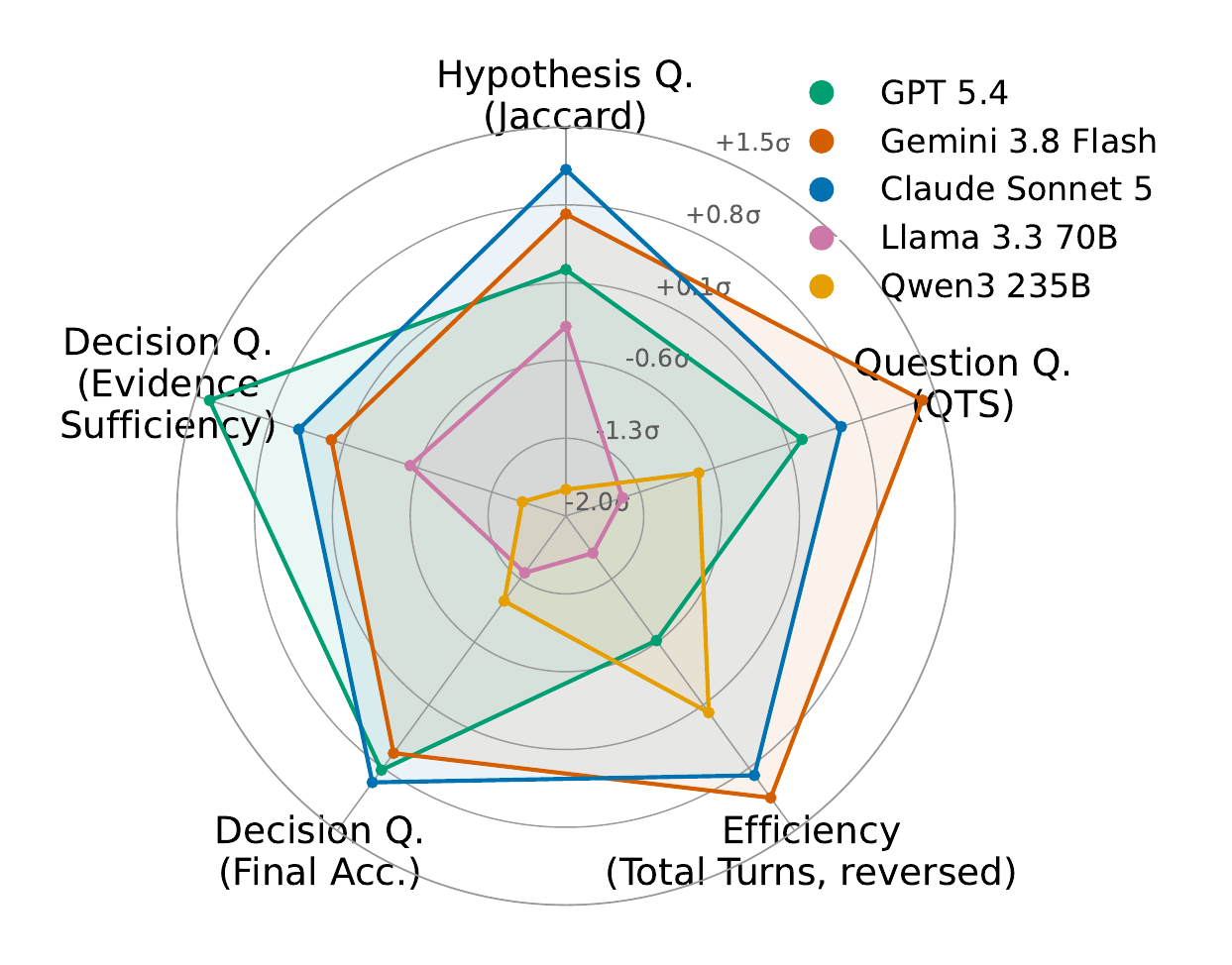}
    \caption{\label{fig:overall_profile}Diagnostic behavior of doctor models under the GPT 5.6 Terra setting across five metrics. Scores are normalized across models. Efficiency is inverted so that higher values indicate fewer turns and better performance.}
    \vspace{-0.2in}
\end{figure}

\paragraph{Distinct Diagnostic Behavior Beyond Final Accuracy.}
\label{sec:result_by_models}

\autoref{fig:overall_profile} and \autoref{tab:main_table} show substantial variation across hypothesis quality, question quality, efficiency, and decision quality. For readability, we report representative values from the GPT 5.6 Terra setting below; the overall behavioral patterns are consistent across both evaluation settings.

\noindent
\textbf{Claude Sonnet 5 maintains selective diagnostic hypotheses.}
It achieves the highest hypothesis precision (0.72) and strong Jaccard scores (0.48), but relatively lower recall (0.59). This suggests that Claude maintains a relatively focused candidate set, excluding more implausible candidates while retaining fewer plausible alternatives. Despite this narrower hypothesis set, it achieves the highest final accuracy (75.94\%).

\noindent
\textbf{Gemini 3.8 Flash combines strong question quality with efficiency.}
It achieves the highest QTS ($0.55$) and requires the fewest turns ($7.30$), while maintaining competitive hypothesis quality and final accuracy. This suggests that Gemini effectively translates its diagnostic hypotheses into targeted follow-up questions with relatively few interactions.

\noindent
\textbf{GPT 5.4 favors hypothesis coverage and evidence gathering.}
It achieves the highest hypothesis recall ($0.65$) and evidence sufficiency ($0.73$), but requires substantially longer interviews ($15.06$ turns). Compared with Claude Sonnet 5, it maintains broader candidate coverage at the cost of lower selectivity and efficiency.

\noindent
\textbf{Llama 3.3 70B struggles to translate hypotheses into effective questioning.}
Its hypothesis quality remains moderate (Jaccard = 0.40), yet it shows the lowest QTS (0.24) and the longest interviews (19.37 turns). This gap suggests that its main bottleneck lies in using diagnostic hypotheses to guide informative follow-up questions.


\noindent
\textbf{Qwen3 235B tends toward earlier diagnostic commitment.}
Qwen3 235B completes interviews faster than Llama 3.3 70B, but shows weaker hypothesis quality and the lowest evidence sufficiency (0.58). This pattern suggests earlier diagnostic commitment with less hypothesis refinement and supporting evidence.

Overall, these results show that diagnostic failures arise at different stages of the interaction.
Models may maintain a broad but less selective differential, form a narrow but effective candidate set, or fail to translate reasonable hypotheses into informative questions.
Our framework makes these distinct process-level behaviors visible beyond final diagnostic accuracy.

\begin{figure}[t]
    \centering
    \includegraphics[width=0.90\linewidth]{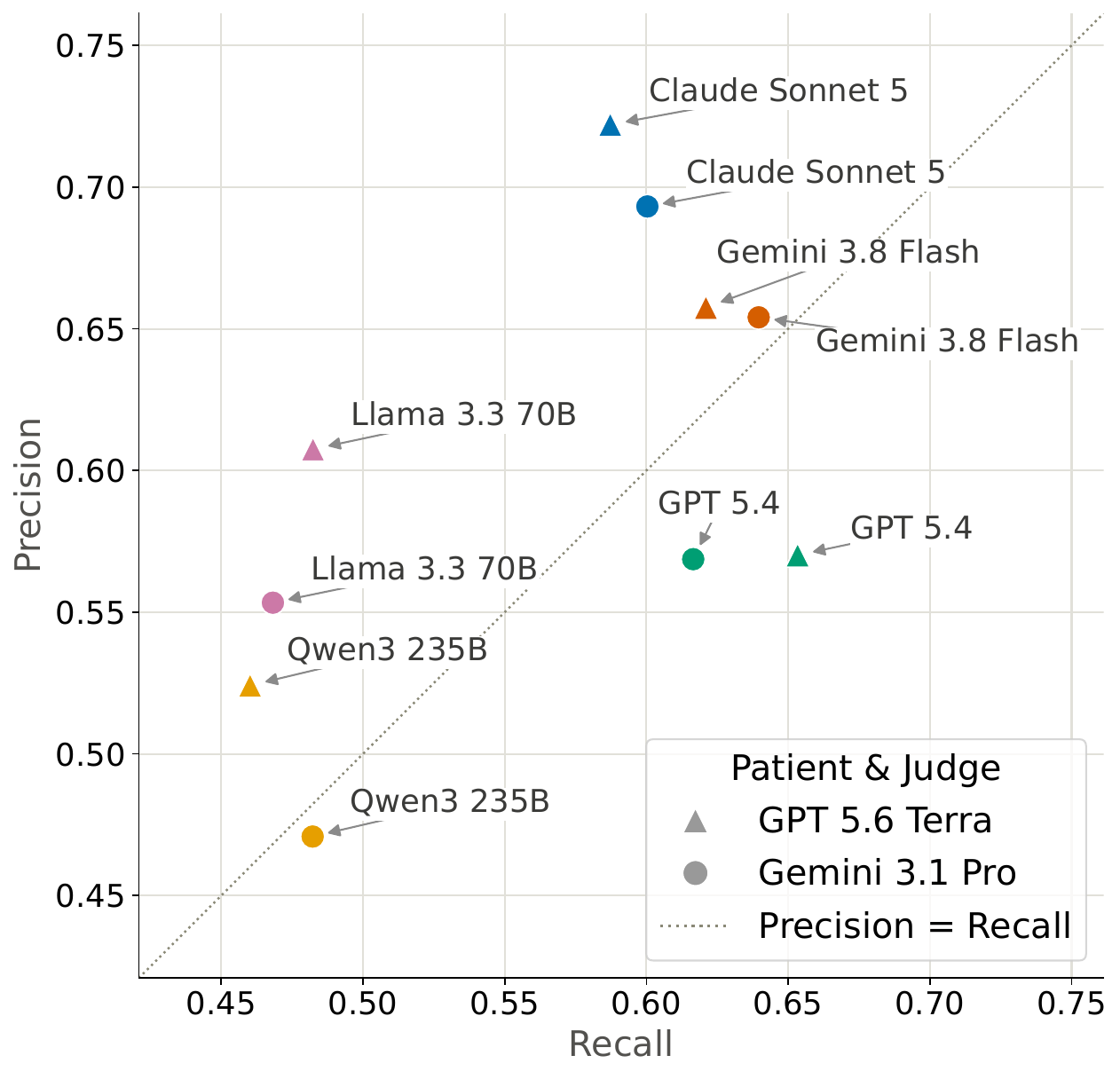}
    \vspace{-0.1in}
    
    \caption{\label{fig:precision_recall}Precision--recall profiles of diagnostic hypothesis formation across doctor models and evaluation settings. The dotted diagonal indicates equal precision and recall.}
    \vspace{-0.1in}
\end{figure}

\subsection{Behavioral Decomposition Analysis}

\paragraph{\dhq{Hypothesis Quality Benefits from Covering Plausible Alternatives.}}
Expert-rated hypothesis quality correlates more strongly with recall ($r=0.44$) than with precision ($r=0.37$). Although both are positively associated with expert judgments, this suggests that good hypothesis quality depends not only on retaining accurate diagnoses, but also on sufficiently covering plausible alternatives. To examine whether models also reflect this emphasis on coverage, \autoref{fig:precision_recall} compares precision and recall under both evaluation settings.

Most models show higher precision than recall across both evaluation settings. This suggests that models tend to focus on narrowing toward a diagnosis while insufficiently considering other disorders that remain plausible given the available evidence. In other words, they may exclude competing diagnoses before gathering enough evidence to rule them out. By evaluating precision and recall throughout the interview, our framework reveals premature exclusion of plausible alternatives that final accuracy alone cannot capture.

\begin{figure}[t]
    \centering
    \includegraphics[width=0.95\linewidth]{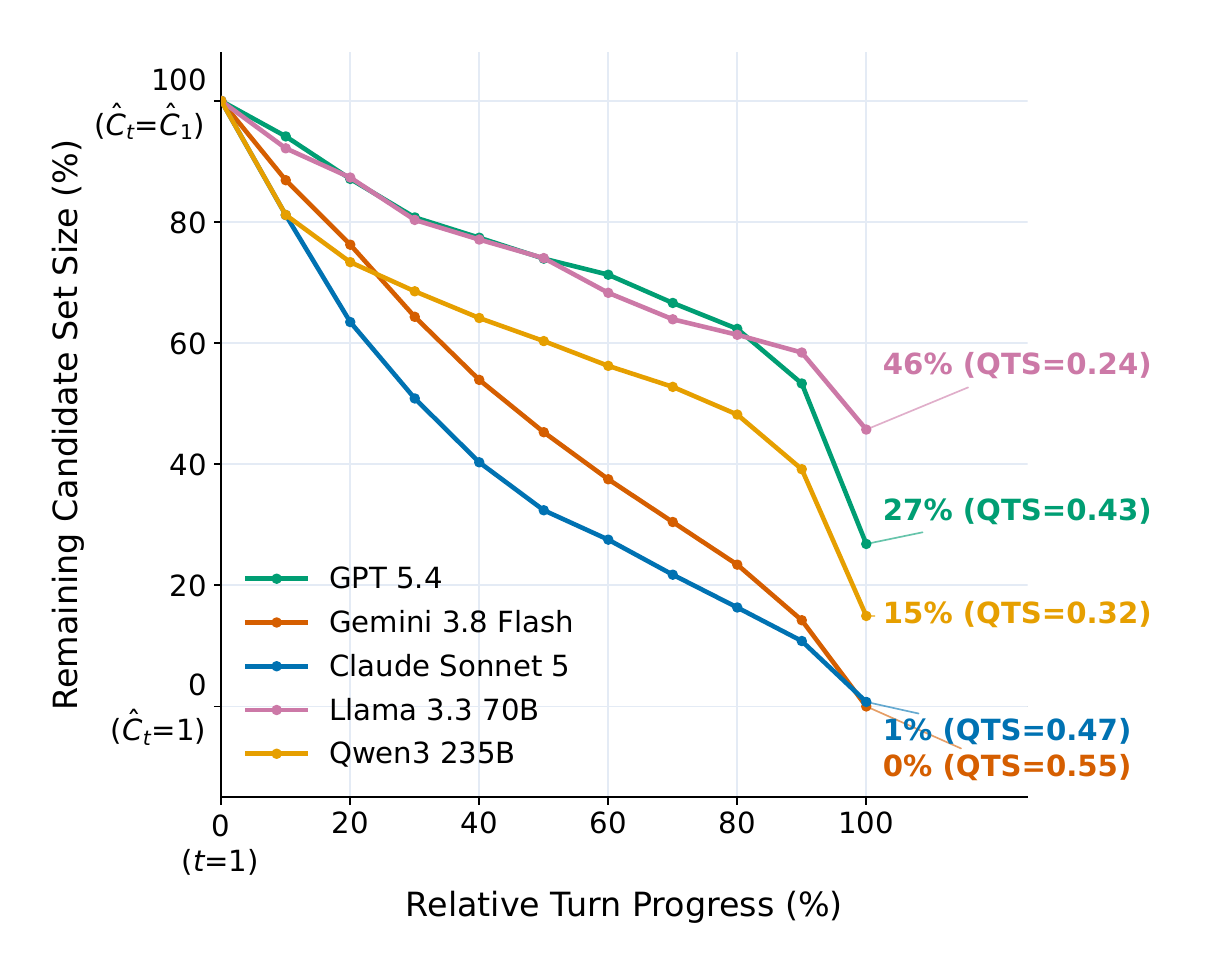}
    \vspace{-0.2in}
    \caption{\label{fig:narrowing progress_qts}Changes in diagnostic candidate-set size across interview progress, together with QTS, under the GPT 5.6 Terra setting. Candidate-set size is shown relative to its initial size.    
    }
    \vspace{-0.1in}
\end{figure}

\paragraph{\dqq{Better Question Targeting Is Associated with Faster Hypothesis Narrowing.}}
Our question-quality evaluation assesses whether the doctor asks questions that target diagnostically relevant and unresolved information. To examine whether better question targeting is associated with more effective narrowing of diagnostic candidates, we analyze the relationship between QTS and candidate-set reduction over dialogue turns. \autoref{fig:narrowing progress_qts} shows how each model's candidate set changes as the interview progresses, with the x-axis representing interview progress and the y-axis representing the remaining candidate-set size.

Models with higher QTS tend to narrow their candidate sets more effectively as the interview progresses. Gemini 3.8 Flash (QTS = 0.55) and Claude Sonnet 5 ($0.47$), for example, progressively reduce their candidate sets and approach a single candidate toward the end of the interview. In contrast, Llama 3.3 70B, which has the lowest QTS ($0.24$), narrows much more slowly and retains a broader candidate set even in later turns. This pattern suggests that effective question targeting is associated with more efficient refinement of the differential diagnosis.

\paragraph{\de{Early Hypothesis Quality Is Linked to Downstream Efficiency.}}

To examine whether early diagnostic quality is associated with downstream interview efficiency, we regress the number of remaining turns after the first three turns on each early-process metric:
\[
R_i = \beta_0 + \beta_1 X_i
+ \alpha_{\mathrm{doctor}(i)}
+ \gamma_{\mathrm{setting}(i)} + \epsilon_i,
\]
where $R_i = T_i - 3$ denotes the number of turns remaining after Turn 3, and $X_i$ denotes Early Jaccard, Precision, Recall, or QTS, averaged over the first three turns. We estimate the model separately for correct and incorrect episodes, controlling for doctor model and evaluation setting. For visualization, interviews are grouped into quartiles based on Early Jaccard and Early Recall, with Q1 representing the lowest values and Q4 the highest.

\begin{figure}[t]
    \centering
    \includegraphics[width=0.95\linewidth]{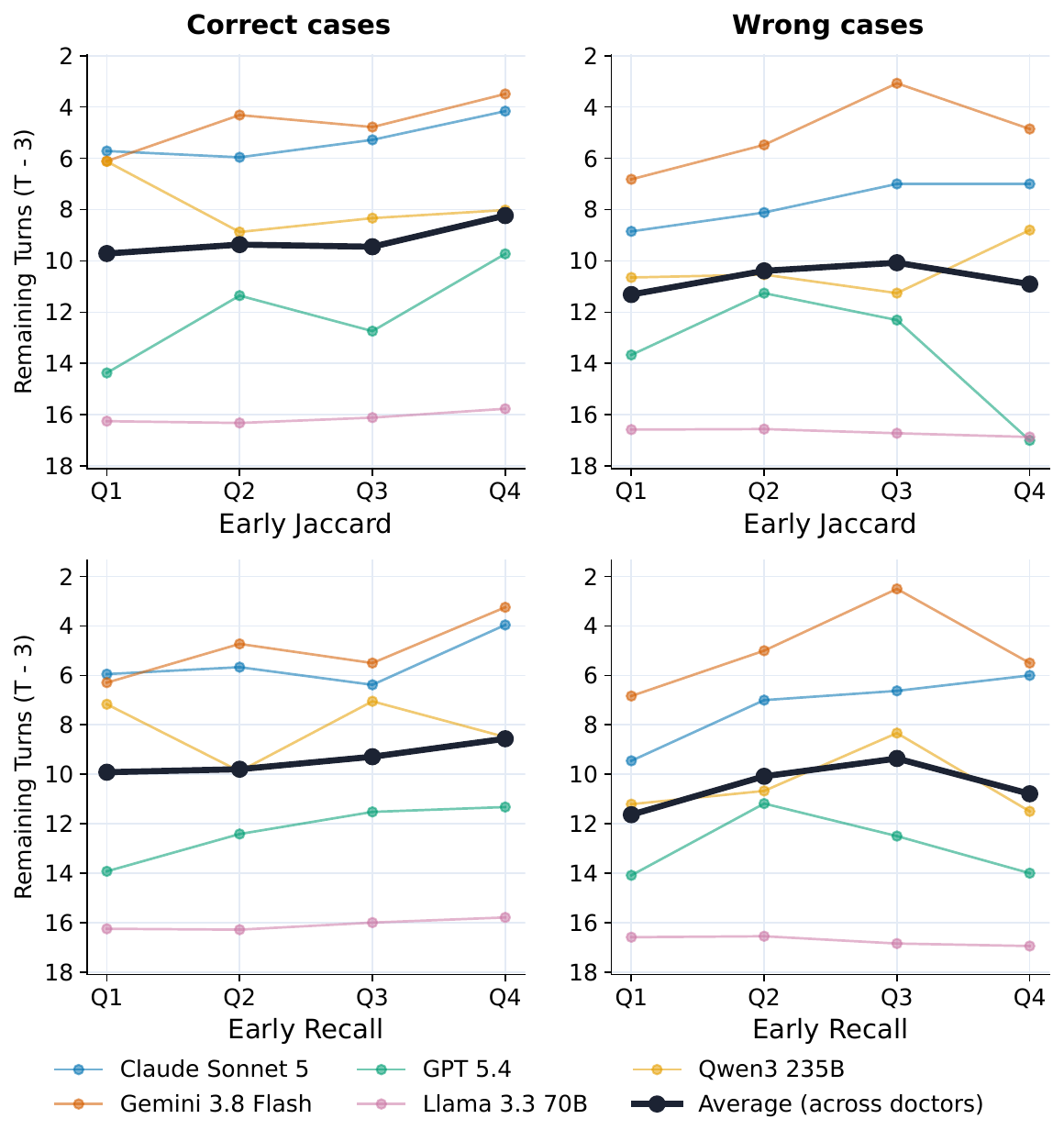}
    \vspace{-0.1in}
    \caption{Early hypothesis quality and downstream interview efficiency. Interviews are grouped into quartiles of Early Jaccard and Recall over the first three turns, with Q1 indicating the lowest values and Q4 the highest, separately for correct and incorrect diagnoses.
    }
    \label{fig:efficiency}
    \vspace{-0.1in}
\end{figure}

As shown in \autoref{fig:efficiency}, higher Early Jaccard and Recall are associated with shorter subsequent interviews, particularly among episodes that ultimately reach the correct diagnosis. After controlling for doctor model and evaluation setting, higher Early Jaccard and Recall are associated with fewer remaining turns in correct episodes ($\beta_1=-2.67$, $p<0.001$ and $\beta_1=-2.61$, $p<0.001$, respectively). Early Precision and QTS show no clear association with subsequent interview length. In incorrect episodes, interview length decreases from Q1 to Q3 but increases again in Q4, yielding no clear monotonic relationship overall.

These results suggest that diagnostic stages are coupled rather than independent. Among interviews that ultimately succeed, establishing an adequate early differential, particularly sufficient coverage of plausible candidates, is associated with a shorter subsequent path to diagnostic resolution.

\begin{figure}[t]
    \centering
    \includegraphics[width=\linewidth]{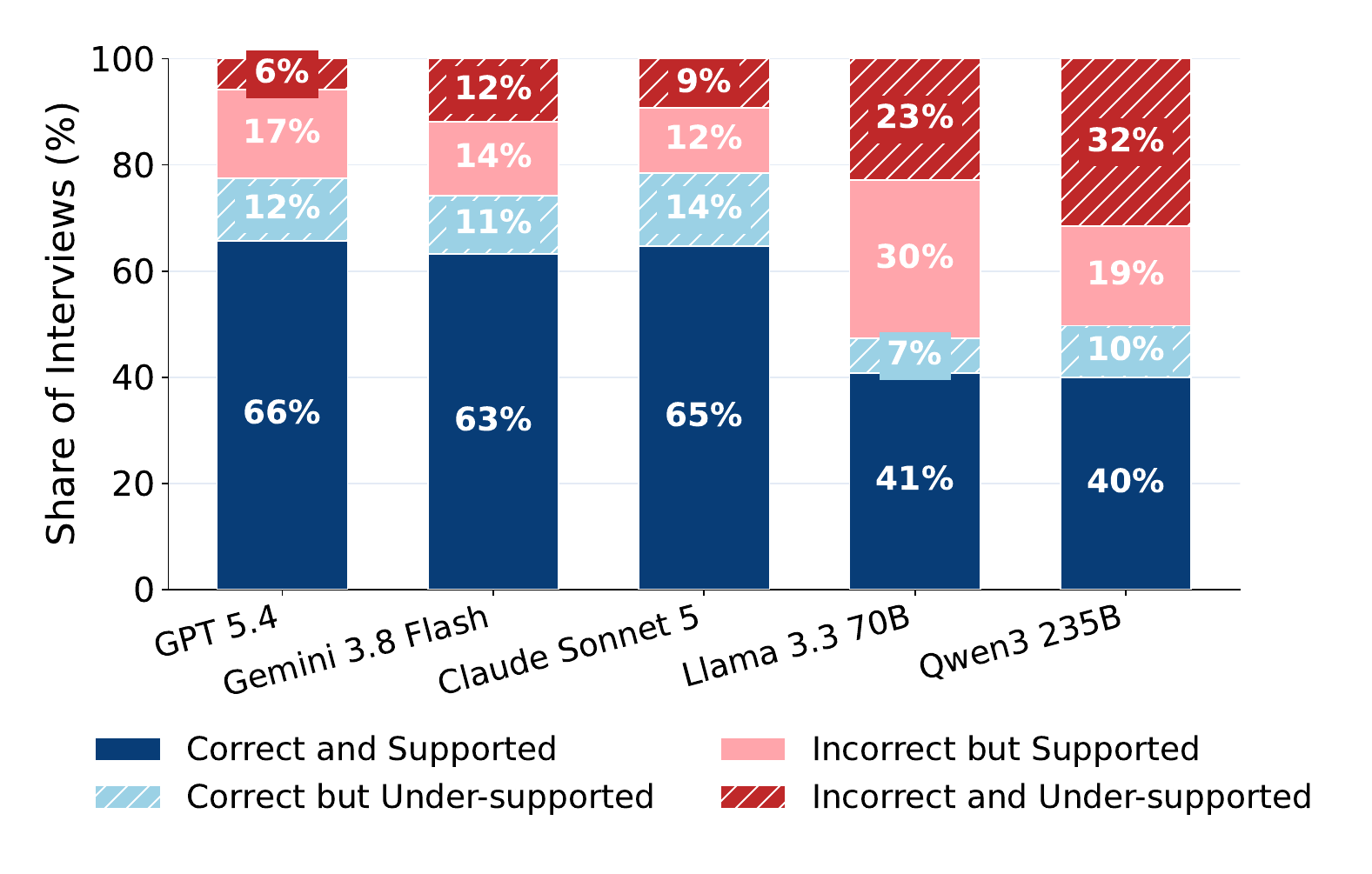}
    \vspace{-0.2in}
    \caption{\label{fig:accuracy_vs_reasoning}Final diagnostic accuracy and evidence sufficiency across doctor models. Interviews are grouped by whether the final diagnosis is correct and sufficiently supported by the acquired evidence.
    }
    \vspace{-0.2in}
\end{figure}

\paragraph{\ddq{Correct Diagnoses Are Not Always Sufficiently Supported.}}
\label{sec:correct_not_supported}
Final Accuracy measures whether the model reaches the correct diagnosis, whereas Diagnostic Evidence Sufficiency (DES) measures whether the acquired evidence sufficiently supports the model's own final diagnosis.
\autoref{fig:accuracy_vs_reasoning} examines these two outcomes jointly. Correct diagnoses are generally accompanied by sufficient evidence, while correct but under-supported decisions are relatively uncommon. However, a substantial fraction of incorrect diagnoses are still supported by the acquired evidence. 
Qwen3 235B and Llama 3.3 70B produce incorrect but supported diagnoses in $19\%$ and $30\%$ of interviews, respectively.

These cases reveal a distinct failure mode in which a model acquires evidence that supports its working diagnosis but ultimately commits to the wrong disorder. Conversely, correct but under-supported diagnoses show that the right answer can be reached without sufficient evidence. By separating diagnostic correctness from evidence sufficiency, our framework distinguishes failures in information acquisition from failures in final diagnostic decision-making.

\section{Conclusion}


In this paper, we introduce \ours{}, a framework for evaluating diagnostic information acquisition (DIA) in multi-turn psychiatric interviews. \ours{} combines a patient simulator that discloses evidence only in response to the doctor's questions with process-level metrics that assess hypothesis quality, question quality, diagnostic efficiency, and final decision quality against the evidence available at each turn, allowing fine-grained analysis of where models succeed or fail along the diagnostic trajectory.
Our analyses show that models with similar final accuracy exhibit distinct diagnostic behaviors in how they form hypotheses, select questions, and gather evidence, and that correct diagnoses are not always supported by sufficient diagnostic evidence.
These findings suggest that LLMs for psychiatric diagnosis should be developed not only to reach the correct diagnosis but also to ground their diagnostic process in adequately acquired evidence. We expect \ours{} to serve as an evaluation framework for assessing LLMs as reliable diagnostic support tools in psychiatry.
\section*{Limitations}

\ours{} evaluates diagnostic information acquisition in controlled simulated interviews rather than real clinical interactions. Although the patient simulator is grounded in clinical profiles derived from the DSM and evaluated for realism and fidelity, simulated interactions cannot fully capture the complexity and variability of real psychiatric encounters. Furthermore, the evaluation is limited to the disorders and diagnostic requirements represented in MentalKG, which may restrict generalization beyond this scope.

The evaluation scope of \ours{} is limited to diagnostic information acquisition (DIA). Since our goal is to assess which evidence a doctor LLM elicits through its questions and how it uses that evidence to reach a diagnosis, we evaluate its diagnostic hypotheses, follow-up questions, efficiency, and final decision against the evidence available at each turn. Accordingly, we do not assess interaction skills such as empathy, rapport building, and safety assessment, which can shape how much a patient discloses in real clinical encounters. Future work could integrate such interaction-level assessments into \ours{} toward a more comprehensive evaluation of psychiatric interviews.

In addition, parts of the evaluation rely on extraction and judgment performed by LLMs. Although we validate the automated evaluation against expert ratings and employ two evaluation settings, the results may still be affected by the choice of evaluation models and extraction procedures. Future work should extend DIPSY to broader clinical conditions, more diverse patient presentations, and real or expert-annotated psychiatric interviews to further assess its generalizability.

\section*{Ethical Consideration}
We address the following ethical considerations in this work. First, all patient profiles and interviews in \ours{} are synthetic and contain no real patient data or personally identifiable information. Some simulated patients describe sensitive experiences such as suicidal ideation or traumatic events, as these are part of the diagnostic criteria of the disorders covered; this content is generated solely for evaluation and is not intended for interaction with real patients or vulnerable users. 
Second, five clinical experts participated in our evaluations: three clinical psychologists for the patient simulator and two psychiatrists for the process metrics. The expert evaluation was approved by the Institutional Review Board (IRB) of our institution (approval number withheld for anonymity). Finally, \ours{} is not intended to promote the use of LLMs as autonomous diagnostic tools. It is designed to evaluate whether LLMs that may assist in mental health assessment ground their diagnostic process in the evidence they acquire. The doctor LLMs evaluated in this work should not be regarded as substitutes for licensed clinicians.

\section*{Acknowledgements}
We thank Chanbi Park (Dongguk University Medical Center) and Gabi Hwang (Chungnam National University Hospital) for their time, clinical expertise, and valuable feedback, which contributed to the development and evaluation of our framework.
We used AI assistants, including ChatGPT\footnote{\url{https://chatgpt.com/}}, Claude\footnote{\url{https://claude.ai/}} and Grammarly\footnote{\url{https://app.grammarly.com/}}, to support the writing and coding processes.


\normalem 
\bibliography{custom} 

\clearpage

\appendix

\section{Implementation for Patient Simulator}
\label{app:patient_simulator}

\begin{table}[ht!]
    \centering
    \small
    \begin{tabular}{@{}lp{0.78\linewidth}@{}}
    \toprule
        \textbf{Code} & \textbf{Disorder} \\
    \midrule
        \multicolumn{2}{@{}l}{\textit{Neurodevelopmental}} \\
        D001 & ADHD, Combined Presentation \\
        D002 & ADHD, Predominantly Inattentive Presentation \\
        D003 & ADHD, Predominantly Hyperactive/Impulsive Presentation \\
    \addlinespace
        \multicolumn{2}{@{}l}{\textit{Schizophrenia Spectrum and Other Psychotic}} \\
        D004 & Delusional Disorder \\
        D005 & Schizophrenia \\
        D006 & Schizoaffective Disorder, Bipolar Type \\
        D007 & Schizoaffective Disorder, Depressive Type \\
    \addlinespace
        \multicolumn{2}{@{}l}{\textit{Bipolar and Related}} \\
        D008 & Bipolar I Disorder \\
        D009 & Bipolar II Disorder \\
        D010 & Bipolar I Disorder with Psychotic Features \\
    \addlinespace
        \multicolumn{2}{@{}l}{\textit{Anxiety}} \\
        D011 & Generalized Anxiety Disorder (GAD) \\
        D012 & Specific Phobia \\
    \addlinespace
        \multicolumn{2}{@{}l}{\textit{Depressive}} \\
        D013 & Major Depressive Disorder (MDD) \\
        D014 & Persistent Depressive Disorder (PDD) \\
        D015 & Major Depressive Disorder with Psychotic Features \\
    \addlinespace
        \multicolumn{2}{@{}l}{\textit{Obsessive-Compulsive and Related}} \\
        D016 & Obsessive-Compulsive Disorder (OCD) \\
        D017 & Body Dysmorphic Disorder \\
    \addlinespace
        \multicolumn{2}{@{}l}{\textit{Trauma- and Stressor-Related}} \\
        D018 & Posttraumatic Stress Disorder (PTSD) \\
        D019 & Acute Stress Disorder \\
        D020 & Adjustment Disorder \\
    \addlinespace
        \multicolumn{2}{@{}l}{\textit{Feeding and Eating}} \\
        D021 & Anorexia Nervosa \\
        D022 & Bulimia Nervosa \\
        D023 & Binge-Eating Disorder \\
    \bottomrule
    \end{tabular}
    \caption{The 23 psychiatric disorders covered in this work, grouped by DSM-5 diagnostic category. Codes follow MentalKG.}
    \label{tab:disease_list}
\end{table}

\subsection{Implementation Details}
\label{app:patient_implementation}
This section provides further details on the patient simulator introduced in Section~\ref{sec:patient_simulator}. We first describe the patient simulator implementation, along with the corresponding prompts (\S\ref{app:patient_implementation}). We then present the quality evaluation, including an analysis of symptom expression diversity (\S\ref{app:patient_quality}), followed by case studies (\S\ref{app:patient_case}).

\subsubsection{Clinical Profile}

Our clinical profiles are drawn from the Type~1 seeds of MentalBench~\citep{song2026mentalbench}. MentalBench samples diagnostic evidence from MentalKG to construct four types of seeds, among which Type~1 seeds are designed to fully satisfy the DSM-5 criteria of a single target disorder: mandatory symptoms are always included, the number of symptoms is sampled at or above the diagnostic threshold, and duration, functional impairment, and additional requirements are explicitly specified. We adopt Type~1 seeds because evaluating how well a doctor model elicits diagnostic evidence requires that the complete evidence be present in the patient's hidden clinical state. These seeds, reviewed by two experts during the construction of MentalBench, comprise five profiles for each of 23 disorders, yielding 115 profiles in total. Table~\ref{tab:disease_list} lists the 23 disorders covered in this work, and Table~\ref{tab:profile_example} shows an example clinical profile.

\begin{table*}[t]
    \centering
    \scriptsize

    \begin{tcolorbox}[
        title={Clinical Profile of Generalized Anxiety Disorder},
        width=0.95\textwidth,
        colback={rgb,255:red,247;green,245;blue,240},
        colbacktitle={rgb,255:red,81;green,78;blue,72},
        coltitle=white,
        colframe=black!20,
        boxrule=0.3pt,
        arc=1pt
    ]

\begin{Verbatim}[
    fontsize=\scriptsize,
    breaklines=true,
    breakanywhere=true
]
{
  "0": {
    "question": {
      "disease_code": "D011",
      "disease_name": "Generalized Anxiety Disorder",
      "sampled_features": {
        "disease_code": "D011",
        "disease_name": "Generalized Anxiety Disorder",
        "sampled_diagnostic_criteria": {
          "functional_impairment_required": true,
          "duration": "9 months"
        },
        "sampled_symptom_groups": {
          "anxiety_symptoms": {
            "symptoms": ["S057", "S056", "S058", "S059", "S027"],
            "name": "anxiety_symptoms"
          }
        },
        "sampled_symptoms": ["S027", "S056", "S057", "S058", "S059"],
        "sampled_symptoms_descriptions": {
          "S027": {
            "name": "Sleep_Disturbance",
            "category": "depressive_episode",
            "description": "A state where there is a significant change in the quantity or quality of sleep.",
            "subtypes": {
              "Initial Insomnia": "Difficulty falling asleep and taking a long time to do so.",
              "Middle Insomnia": "Waking up frequently during sleep or having difficulty falling back asleep.",
              "Early Morning Awakening": "Waking up too early and being unable to fall back asleep.",
              "Hypersomnia": "Sleeping for long periods at night or excessive sleep during the day.",
              "Nightmare Disorder": "Sleep disturbance caused by repeated and distressing dreams.",
              "Non-Restorative Sleep": "Feeling unrefreshed despite having slept enough."
            },
            "examples": null
          },
          "S056": {
            "name": "Excessive_Anxiety_Worry",
            "category": "anxiety_symptoms",
            "description": "Excessive anxiety and worry (expectant apprehension) occurring persistently about a number of events or activities.",
            "subtypes": {
              "Chronic Worry": "Worry occurs more days than not for at least 6 months.",
              "Generalized Worry": "Worry is not confined to specific topics but spreads to various domains like work, school, family, and finances.",
              "Catastrophizing": "Jumping to conclusions that minor mistakes or events will lead to irreversible disasters.",
              "Intolerance of Uncertainty": "Unable to tolerate future uncertainty, trapped in 'what if' questions."
            },
            "examples": null
          },
          ...
        }
      }
    }
  }
}
\end{Verbatim}

    \end{tcolorbox}

    \caption{Example of a sampled clinical profile for Generalized Anxiety Disorder.}
    \label{tab:profile_example}
\end{table*}

\subsubsection{Context-Grounded Symptom Manifestation}
We ground each profile in a concrete patient context through the following procedure. All generation steps use GPT 5.5 with a temperature of 0.9.

\paragraph{Onset narrative.} Each profile is instantiated as three patients with distinct personas sampled from Nemotron-Personas-USA~\citep{nvidia/Nemotron-Personas-USA}. Each persona is assigned the life event, among 32 events from the Holmes--Rahe Social Readjustment Rating Scale~\citep{holmes1967social}, that is most plausible given the persona's age, occupation, and living situation, with the three patients of the same profile receiving different events. The three patients are also randomly assigned one of mild, moderate, and severe severity levels, each used once. Conditioned on these, we generate a three- to four-sentence onset narrative describing how early signs progressed to a clinical level over the specified duration and how the patient came to seek care (Table~\ref{tab:prompt_stressor_generation}).

\paragraph{Manifestation.} For each symptom in the profile, we then generate a single first-person sentence describing how the patient experiences it in daily life. The sentence avoids naming the symptom or using diagnostic terminology, matches its vocabulary and complexity to the patient's education and background, and, together with the other manifestations, reflects the assigned severity. Symptom subtypes from MentalKG are not sampled but provided only as a reference for expression (Table~\ref{tab:prompt_manifestation_generation}). Duration, functional impairment, and additional requirements are expressed in the same manner in one to two sentences each, with functional impairment described as real-life consequences rather than a restatement of symptoms, and symptom-group durations kept consistent with the overall duration (Table~\ref{tab:prompt_additional_requirement_generation}).

\paragraph{Chief complaint.} From the 84 symptoms in MentalKG, experts identified 25 that patients are unlikely to present as their reason for visiting (e.g., inflated self-esteem in a manic episode) and defined them as an exclusion list. The chief complaint is randomly selected from the profile's symptoms that are not on this list, and its manifestation serves as the patient's first-turn utterance.

\subsubsection{Controlled Information Disclosure}
Each patient response is generated in two stages, alignment and generation, both using GPT 5.6 Terra or Gemini 3.1 Pro depending on the evaluation setting (temperature 0.7, top-$p$ 0.9). Beyond making each response traceable to the question that elicited it, this design structurally prevents the patient from fabricating clinical information absent from the profile, rather than relying on prompt instructions alone.

\paragraph{Alignment.} Given the list of profile manifestations, the patient's demographics and chief complaint symptom code, the dialogue history, and the doctor's latest question, this stage returns the \emph{single} profile item that the question targets. When a question touches on several topics, the item it most directly addresses is selected; when the profile contains no matching item, \textit{none} is returned instead of forcing an unrelated match. In the first turn, the chief complaint item is always returned regardless of the doctor's opening question (Table~\ref{tab:prompt_profile_alignment}). Since the doctor model is also instructed to ask about one thing at a time (Table~\ref{tab:prompt_question_generation}), each turn pairs one question with one disclosed item.

\paragraph{Generation.} This stage receives only the dialogue history, the latest question, the patient's demographics and severity, and the aligned item with its manifestation. The onset narrative is not provided at this stage, so the patient's background enters the dialogue only through the pre-generated manifestations. The patient responds as in natural speech, with hesitations and self-corrections, and when the same item is selected again, it conveys a different aspect or acknowledges having mentioned it rather than repeating the earlier answer. The patient is also instructed not to confirm specific details named by the doctor unless they are contained in the aligned item. When \textit{none} is returned, the patient responds without new diagnostic information, for example by expressing uncertainty or asking for clarification, while varying its phrasing (Table~\ref{tab:prompt_patient_response_generation}).

\begin{table*}[t]
    \centering
    \scriptsize

    \begin{tcolorbox}[
        patientprompt,
        title={Prompt for Stressor Generation},
        width=\textwidth
    ]
    \ttfamily
    \raggedright

You are a psychiatric case scenario writer. Combine the (A) diagnostic profile and (B) real personas below to write, in English, the life background and circumstances surrounding the onset, worsening, recognition, or presentation of this disorder for each persona - what happened and how the condition became clinically significant.

\par\smallskip

The goal is a coherent background story, not a description of the symptoms themselves. Symptom-level detail will be written separately in a later stage, so keep the stressor focused on background rather than symptom description.

\medskip
\textbf{[A. Diagnostic Profile - Shared by Every Persona Below]}

- Disorder: \{disease\_name\} (\{disease\_code\})\\
- Symptom group: \{symptom\_group\_name\} (duration: \{duration\}, functional impairment required: \{functional\_impairment\_required\})\\
- Symptoms this person has - the background must be built on these: \{symptom\_list\_with\_names\}

\medskip
\textbf{[B. Personas - Write One Independent Result per Persona, in the Same Order]}

--- Persona \{index\} (uuid: \{uuid\}) ---\\
- Sex / Age: \{sex\} / \{age\}\\
- Marital status: \{marital\_status\}\\
- Education: \{education\_level\}\\
- Occupation: \{occupation\}\\
- General persona: \{persona\}\\
- Professional background: \{professional\_persona\}\\
- Cultural background: \{cultural\_background\}\\
- Hobbies / interests: \{hobbies\_and\_interests\}\\
- Sports context: \{sports\_persona\}\\
- Arts context: \{arts\_persona\}\\
- Travel context: \{travel\_persona\}\\
- Culinary context: \{culinary\_persona\}

\medskip
\textbf{[Decision Priorities - Apply in This Exact Order]}

1. Clinical plausibility: The relationship between the disorder and the life background must be clinically coherent. Do not assume that every disorder was directly caused by a single external event; represent the relationship between the circumstances and the onset, worsening, recognition, or presentation of the condition accurately.

\par\smallskip

2. Biographical coherence: The background must fit the persona's age, relationships, cultural and social environment, interests, daily life, education, and professional circumstances. Use these details to shape what matters to the person, how the circumstances affect them, and why they seek care.

\par\smallskip

3. Diversity across the three cases: When multiple backgrounds are similarly plausible, prefer meaningfully different life domains, events, and clinical pathways across the personas. Repetition is acceptable when it is clearly the most plausible choice.

\par\smallskip

4. Creativity: Make each background concrete, individualized, and memorable while remaining realistic. Creative details must never override clinical or biographical plausibility.

\medskip
\textbf{[Writing Guidelines]}

1. First, fill in \texttt{disease\_info}: In 2-3 sentences, summarize \{disease\_name\} as a psychiatrist would already know it before seeing a patient - its defining diagnostic features, typical presentation and course, and what distinguishes it from similar disorders.

\par\smallskip

2. For EACH of the \{num\_personas\} personas above, fill in exactly three fields: \texttt{stressor\_event}, \texttt{severity}, \texttt{stressor}.

\par\smallskip

3. \texttt{stressor\_event}: Each persona's background must be built on ONE life event taken from the candidate list below. Pick the event that this specific persona's age, marital status, occupation, and living situation make most plausible - never one their circumstances rule out. The \{num\_personas\} personas must each be given a DIFFERENT event. Copy the chosen event into \texttt{stressor\_event} exactly as it is written in the list.

\par\smallskip

Candidate life events: \{stressor\_event\_candidates\}

\par\smallskip

4. \texttt{severity}: Assign each persona one of Mild, Moderate, or Severe at random, using each of the three levels exactly once across the \{num\_personas\} personas.

\par\smallskip

5. \texttt{stressor}: In 3-4 sentences, write this persona's disease course as a single arc, following these principles:

(1) Imagine this persona at the moment they went through the assigned \texttt{stressor\_event}.

(2) Starting from the earliest signs, describe how those grew to a clinical level over the \{duration\}, in terms of the symptoms and diagnostic requirements listed above.

(3) End with how this person came to receive psychiatric care.

\par\smallskip

Cautions:

- Build the narrative only on the symptoms and requirements given in the profile.\\
- Vary the route into care across the personas so it is not always someone else urging them.\\
- Keep the narrative consistent with the assigned severity.

\par\smallskip

6. Each persona's narrative must stay fully consistent with the shared diagnostic profile, and must not introduce conditions, episodes, or events the profile does not contain (e.g., manic episodes when no manic symptoms are present).

\medskip
\textbf{[Output Format]}

Output JSON only, no other text. \texttt{"results"} must have exactly \{num\_personas\} items, in the same order as the personas listed above.

\begin{verbatim}
{
  "disease_info": "...",
  "results": [
    {
      "stressor_event": "...",
      "severity": "Mild | Moderate | Severe",
      "stressor": "..."
    }
  ]
}
\end{verbatim}

    \end{tcolorbox}
    \caption{\label{tab:prompt_stressor_generation}Prompt for Stressor Generation}
    
\end{table*}
\begin{table*}[t]
    \centering
    \scriptsize

    \begin{tcolorbox}[
        patientprompt,
        title={Prompt for Manifestation Generation},
        width=0.95\textwidth
    ]
    \ttfamily
    \raggedright

You are a mental health expert creating realistic clinical cases for a doctor-patient interview simulator.

\par\smallskip

Using the disease information, symptom profile, and established patient case below, write how this specific patient would actually experience each assigned symptom in daily life. Write one manifestation sentence per symptom, in the patient's own voice, as though they were describing the experience during a psychiatric interview.

\par\smallskip

Do not invent new symptoms or change the established diagnosis, severity, or patient background. Your goal is to create clinically plausible, diverse, and highly individualized manifestations that reflect how a real person with \{disease\_name\} might experience the assigned symptoms.

\medskip
\textbf{[A. Disease Information]}

\{disease\_info\}

\medskip
\textbf{[B. Symptom Profile]}

Disorder: \{disease\_name\} (\{disease\_code\})\\
Duration: \{duration\}\\
Symptoms: \{symptom\_block\}

\medskip
\textbf{[C. Established Patient Case]}

Demographics: \{demographics\}\\
Stressor event: \{stressor\_event\}\\
Overall severity: \{severity\}\\
Stressor: \{stressor\}\\
Chief complaint symptom code: \{chief\_complaint\_symptom\_code\}

\medskip
\textbf{[Writing Guidelines]}

1. For each symptom code, write exactly ONE sentence describing its manifestation for this patient. Where a symptom lists reference subtypes/example presentations, use them as inspiration for a concrete, specific angle - you don't need to pick or report exactly one, and don't copy any example verbatim.

\par\smallskip

2. Write in the first person and in the patient's natural speaking voice. The sentence should sound like something the patient would actually report to a psychiatrist, not a clinical case description.

\par\smallskip

3. Describe the patient's concrete lived experience rather than naming the symptom, repeating diagnostic criteria, or using psychiatric terminology.

\par\smallskip

4. Make each manifestation specific to this individual by drawing naturally on the established demographics, stressor, severity, and daily environment. Avoid generic descriptions that could apply to any patient. Use the stressor only when it is clinically relevant to the specific symptom. Do not force every manifestation to explicitly mention or directly result from the stressor.

\par\smallskip

5. Match the vocabulary, sentence complexity, and manner of expression to the patient's education level and background.

\par\smallskip

6. The manifestations as a whole should reflect the assigned overall severity (\{severity\}). Individual symptoms may vary in intensity, prominence, and functional impact.

\par\smallskip

7. Reflect that the condition has persisted for \{duration\} rather than describing a single isolated event. Express the persistence naturally and do not force the exact duration into every sentence.

\par\smallskip

8. The chief-complaint symptom is what the patient opens the interview with, so write that one manifestation as the sentence they would volunteer first when asked what brings them in.

\par\smallskip

9. Taken together, all manifestations must sound like experiences reported by one coherent, realistic patient with \{disease\_name\}, without contradicting the disease information or established patient case.

\medskip
\textbf{[Output Format]}

Output JSON only, no other text. Include exactly one key for every symptom code listed above. Return exactly the symptom codes listed in the prompt. Do not omit, rename, or add any symptom code.

\begin{verbatim}
{
  "<symptom_code>": "<manifestation sentence>"
}
\end{verbatim}

    \end{tcolorbox}
    \caption{\label{tab:prompt_manifestation_generation}Prompt for Manifestation Generation}
    
\end{table*}

\begin{table*}[t]
    \centering
    \scriptsize

    \begin{tcolorbox}[
        patientprompt,
        title={Prompt for Additional Requirement Generation},
        width=0.95\textwidth
    ]
    \ttfamily
    \raggedright

You are a mental health expert creating realistic clinical cases for a doctor-patient interview simulator.

\par\smallskip

The patient's symptom manifestations have already been written. Your task is to generate brief manifestation-style evidence for information that is not captured by the symptom manifestations, including:

(C) additional diagnostic requirements\\
(D) functional impairment\\
(E) duration-related evidence

\medskip
\textbf{[A. Disease Information]}

\{disease\_info\}

\par\smallskip

Disorder: \{disease\_name\} (\{disease\_code\})

\medskip
\textbf{[B. Established Patient Case]}

Demographics: \{demographics\} \\
Stressor event: \{stressor\_event\} \\
Overall severity: \{severity\} \\
Stressor: \{stressor\} \\
Chief complaint symptom code: \{chief\_complaint\_symptom\_code\}

\medskip
\textbf{[B2. Established Symptom Manifestations]}

Reference only.

\par\smallskip

\{manifestation\_block\}

\medskip
\textbf{[C. Additional Diagnostic Requirements]}

\{additional\_requirements\_block\}

\medskip
\textbf{[D. Functional Impairment]}

Functional impairment required:\\ \{functional\_impairment\_required\}

\medskip
\textbf{[E. Duration Information]}

\{duration\_items\_block\}

\medskip
\textbf{[Writing Guidelines]}

1. Generate ONE brief manifestation, 1-2 sentences maximum, for each item.

\par\smallskip

2. Use the patient's first-person voice throughout.

\par\smallskip

3. Additional diagnostic requirements:\\
- Write concrete evidence showing that the requirement is satisfied.\\
- Ground it in the patient's own history or lived experience.\\
- Do not simply restate the criterion.\\
- If there are no additional requirements, return an empty object.

\par\smallskip

4. Functional impairment: \\
- Describe the real-life consequence of the accumulated symptoms rather than repeating the symptoms themselves.\\
- Focus on impairment in occupational, academic, social, self-care, or safety domains.\\
- Keep it brief, 1-2 sentences.

\par\smallskip

5. Duration:\\
There are two kinds of duration:\\
- \texttt{duration}: Refers to the overall duration of the disorder.\\
- \texttt{<symptom\_group>\_duration}: Refers only to how long that specific symptom group has been present within the overall illness. \\
\par\smallskip
For every duration item: \\
- Write one brief first-person sentence.\\
- It is sufficient to mention that the symptoms have been present since a memorable life event, season, job change, or approximate period, e.g., ``Ever since the station rebranding...'' or ``It's been going on for about two years now...''\\
- Do not make the duration overly detailed.\\
- Ensure that symptom-group durations are consistent with the overall duration.

\par\smallskip

6. Keep everything consistent with the patient's demographics, stressor, severity, and existing manifestations.

\par\smallskip

7. Do not introduce new symptoms or contradict the established manifestations.

\medskip
\textbf{[Output Format]}

Output JSON only.

\begin{verbatim}
{
  "additional_requirements": {
    "<requirement>": "<1-2 sentence manifestation>"
  },
  "functional_impairment": "<1-2 sentence manifestation>",
  "<duration key>": "<1-2 sentence manifestation>"
}
\end{verbatim}

    \end{tcolorbox}
    \caption{\label{tab:prompt_additional_requirement_generation}Prompt for Additional Requirement Generation}
    
\end{table*}
\begin{table*}[t]
    \centering
    \scriptsize

    \begin{tcolorbox}[
        patientprompt,
        title={Prompt for Profile Alignment},
        width=0.95\textwidth
    ]
    \ttfamily
    \raggedright

You are a patient who has come to a psychiatric clinic with \{disease\_name\}. You are about to talk with the doctor.

\par\smallskip

Looking at the dialogue so far and the doctor's last question, work out what it is asking about, and pick the item of your symptom profile to answer it with.

\medskip
\textbf{[Symptom Profiles]}

\{manifestation\}

\medskip
\textbf{[Patient Info]}

\{patient\_info\}

\medskip
\textbf{[Dialogue History]}

\{dialogue\_history\}

\medskip
\textbf{[Last Question]}

\{last\_question\}

\medskip
\textbf{[Rules]}

\begin{itemize}[
    label=-,
    leftmargin=1.2em,
    labelsep=0.4em,
    itemsep=0pt,
    topsep=0pt
]
    \item Return exactly ONE item. If the question touches several topics, return the one it is most directly about - never more than one.
    \item Copy \texttt{key}, \texttt{name}, and \texttt{description} exactly as they appear in [Symptom Profiles].
    \item If your profile does not cover what the doctor asked, return the text ``None'' for \texttt{key}, \texttt{name}, and \texttt{description}, and say why in \texttt{reason}. Do not force an unrelated item.
    \item On your first reply of the consultation, always return the item whose key is your \texttt{chief\_complaint\_symptom\_code}, whatever the doctor opened with.
    \item Do not answer the doctor, and do not write anything outside the JSON.
\end{itemize}

\medskip
\textbf{[Output Format]}

Output JSON only, no other text.

\begin{verbatim}
{
  "key": "<key>",
  "name": "<name>",
  "description": "<description>",
  "reason": "<why this item>"
}
\end{verbatim}

    \end{tcolorbox}
    \caption{\label{tab:prompt_profile_alignment}Prompt for profile alignment}
    
\end{table*}
\begin{table*}[t]
    \centering
    \scriptsize

    \begin{tcolorbox}[
        patientprompt,
        title={Prompt for Patient Response Generation},
        width=0.95\textwidth
    ]
    \ttfamily
    \raggedright

\par\smallskip

You are a patient experiencing \{disease\_name\}, in a diagnostic interview with a psychiatrist. Your goal is to answer the doctor's question naturally, using the information you are given.

\par\smallskip

Answer with the [Targeted Info].

\medskip
\textbf{[Dialogue History]}

\{dialogue\_history\}

\medskip
\textbf{[Last Question]}

\{last\_question\}

\medskip
\textbf{[Patient Info]}

\{patient\_info\}

\medskip
\textbf{[Targeted Info]}

\{targeted\_info\}

\medskip
\textbf{[What to Do]}

\begin{itemize}[
    label=-,
    leftmargin=1.2em,
    labelsep=0.4em,
    itemsep=0pt,
    topsep=0pt
]
    \item Answer the doctor's last question with what [Targeted Info] holds, in your own words.
    \item You are speaking out loud, not reading a line you prepared. People restart sentences, pause to think, trail off, and come at the point sideways - let yours do that where it would.
    \item Speak in the first person, and let how you feel about what you are saying come through.
    \item If you are handed the same item again, say a different part of it, or answer as someone who knows they have already said this - not the same sentence twice.
    \item If [Targeted Info] is ``None'', answer as a person who simply does not have that to give: say you are not sure, say you have not noticed it, ask what the doctor means, or say what you do know and stop there. Do not fall back on the same phrase every time.
\end{itemize}

\medskip
\textbf{[What Not to Do]}

\begin{itemize}[
    label=-,
    leftmargin=1.2em,
    labelsep=0.4em,
    itemsep=0pt,
    topsep=0pt
]
    \item Don't bring up symptoms the doctor did not ask about, even ones you have.
    \item Don't state a clinical fact that is not in [Targeted Info] - no other symptoms, medications, test results, dates, or events.
    \item Don't confirm a specific the doctor names if [Targeted Info] does not contain it. Answer about what you were given instead.
    \item Don't deny what [Targeted Info] says. You may minimize it or hold back, never contradict it.
    \item Don't use diagnostic terms or name your disorder.
    \item Don't write anything but your reply, in English - no labels, quotation marks, or narration.
\end{itemize}

    \end{tcolorbox}
    \caption{\label{tab:prompt_patient_response_generation}Prompt for patient response generation}
    
\end{table*}

\subsection{Quality Evaluation}
\label{app:patient_quality}
This section presents the results of evaluating the quality of our patient simulator. We first conduct an expert evaluation, in which clinical psychologists assess the realism and profile fidelity of sampled conversations (\S\ref{app:pateint_expert_eval}). To extend this assessment beyond the expert-rated sample, we then perform an automated evaluation at scale using an LLM judge under the same criteria (\S\ref{app:patient_auto_eval}). Finally, we examine whether our context-grounded patients express symptoms in diverse ways, both within and across disorders (\S\ref{app:patient_symptom_diversity}).

\subsubsection{Expert Evaluation}
\label{app:pateint_expert_eval}

\paragraph{Evaluators.}
The expert evaluation was conducted by three licensed clinical psychologists working at university or general hospitals in South Korea. All evaluators hold a government-issued clinical psychologist license, and one is additionally certified as a clinical psychology specialist by a national professional association. Their qualifications and clinical experience are summarized in Table~\ref{tab:psychologist_info}.

\begin{table}[H]
    \centering
    \small
    \begin{tabular}{@{}cp{0.52\linewidth}c@{}}
    \toprule
        \textbf{Evaluator} & \textbf{Qualification} & \textbf{Exp. (yrs)} \\
    \midrule
        E1 & National license (Grade I); \newline Specialist certification & 7--10 \\
    \addlinespace
        E2 & National license (Grade II) & 1--3 \\
    \addlinespace
        E3 & National license (Grade II) & $<$1 \\
    \bottomrule
    \end{tabular}
    \caption{Qualifications and post-licensure clinical experience (in years) of the expert evaluators. ``National license'' denotes the government-issued clinical psychologist license, and ``Specialist certification'' denotes certification as a clinical psychology specialist by a national professional association.}
    \label{tab:psychologist_info}
\end{table}

\paragraph{Case Sampling.}
We sampled 46 conversations from the main experiments, two for each of the 23 disorders. For each disorder, the two conversations were drawn from different clinical profiles, one simulated with each patient backbone (GPT 5.6 Terra and Gemini 3.1 Pro). The doctor model was randomly selected from Claude Sonnet 5, Gemini 3.8 Flash, and GPT 5.4 (16, 13, and 17 conversations, respectively). We excluded Llama 3.3 70B and Qwen3 235B because they tend to repeat the same questions, leading the patient to repeatedly deny the same content and potentially lowering patient ratings regardless of the simulator's own quality.

\paragraph{Evaluation Materials.}
Each case consisted of the clinical profile and the full interview transcript. The profile presented the target diagnosis, demographics, onset and background, severity, chief complaint, symptom manifestations, and the manifestations of duration, functional impairment, and additional requirements. Doctor utterances in the transcript were shown only to indicate what each patient response was answering, and evaluators were asked to focus solely on the patient agent's utterances. All cases were translated into Korean using GPT 5.4, and evaluators were instructed not to penalize awkwardness attributable to translation.

\paragraph{Evaluation Criteria.}
Evaluators rated each case on three dimensions using a 1--5 scale. \textit{Profile realism} assesses whether each symptom manifestation is phrased as a real patient would describe the symptom. \textit{Conversation realism} assesses whether the patient agent's utterances feel like those of a real patient. \textit{Profile fidelity} assesses whether the patient agent adheres to the given clinical profile; stating clinical information not specified in the profile, contradicting the profile, confirming such information when raised by the doctor, and omitting or replacing the chief complaint were defined as violations. Expressing uncertainty when asked about information not specified in the profile, or not mentioning symptoms the doctor never asked about, was not considered a violation. The scoring rubric is presented in Table~\ref{tab:expert_rubric}.

\textbf{\begin{table*}[t]
    \centering
    \small
    \begin{tabular}{@{}c p{0.29\textwidth} p{0.29\textwidth} p{0.29\textwidth}@{}}
    \toprule
        \textbf{Score} & \textbf{Profile Realism} & \textbf{Conversation Realism} & \textbf{Profile Fidelity} \\
        & \textit{Is each symptom manifestation phrased as a real patient would describe the symptom?}
        & \textit{Do the patient agent's utterances feel like those of a real patient?}
        & \textit{Does the patient agent adhere to the given clinical profile?} \\
    \midrule
        5 & All sentences meet the criterion.
          & Utterances throughout resemble those of a real patient.
          & The patient responds faithfully to the clinical profile. \\
    \addlinespace
        4 & All sentences meet the criterion, but some expressions are slightly awkward.
          & Some expressions are awkward, but utterances overall resemble those of a real patient.
          & One or two minor violations that do not affect the diagnosis. \\
    \addlinespace
        3 & One or two sentences do not meet the criterion.
          & Utterances implausible for a real patient appear occasionally (1--2 turns).
          & Three or more minor violations that do not affect the diagnosis. \\
    \addlinespace
        2 & Three or more sentences do not meet the criterion.
          & Utterances implausible for a real patient appear frequently (3 or more turns).
          & At least one violation that could affect the doctor's diagnostic judgment. \\
    \addlinespace
        1 & Most sentences do not meet the criterion.
          & Utterances throughout are implausible for a real patient.
          & The conversation proceeds largely unrelated to the clinical profile. \\
    \bottomrule
    \end{tabular}
    \caption{Scoring rubric for the expert evaluation of the patient simulator. Each dimension is rated on a 1--5 scale according to the guiding question shown below its name.}
    \label{tab:expert_rubric}
\end{table*}}

\paragraph{Evaluation Results.}
Table~\ref{tab:patient_expert_agreement} reports each expert's mean rating per dimension, the mean of majority-voted ratings, and inter-rater agreement. All three dimensions received generally high ratings, and inter-rater agreement was also high (AC$_2$ $=$ .86 -- .92). E1's Profile Fidelity rating was lower than those of the other two experts, reflecting a difference in interpreting the guideline: E1 regarded the patient's denial of symptoms not specified in the profile as a violation, whereas the other two experts did not. This behavior is partly intended, as the patient is instructed to either express uncertainty or state that it has not noticed such symptoms (Table~\ref{tab:prompt_patient_response_generation}). Because each profile fully satisfies the diagnostic criteria of the target disorder, symptoms not in the profile are treated as absent, and denying them can rule out alternative diagnoses but never the target disorder. We therefore retained the simulator's behavior without modification.
\begin{table}[H]
    \centering
    \resizebox{\columnwidth}{!}{%
    \begin{tabular}{@{}lccccc@{}}
    \toprule
        & \multicolumn{4}{c}{\textbf{Mean Rating}} & \\
        \cmidrule(lr){2-6}
        \textbf{Dimension} & \textbf{E1} & \textbf{E2} & \textbf{E3} & \textbf{Majority} & \textbf{AC$_2$} \\
    \midrule
        Profile Realism      & 4.93 & 4.43 & 4.64 & 4.85 & 0.92 \\
        Conversation Realism & 4.46 & 4.61 & 4.61 & 4.65 & 0.91 \\
        Profile Fidelity     & 3.93 & 4.87 & 4.75 & 4.72 & 0.86 \\
    \bottomrule
    \end{tabular}%
    }
    \caption{Expert ratings of the patient simulator on 46 sampled conversations (1--5 scale). E3 rated 44 of the 46 cases. Majority vote takes the modal score, or the rounded mean when all three experts differ. AC$_2$: Gwet's AC$_2$ with quadratic weights across all three experts.}
    \label{tab:patient_expert_agreement}
\end{table}

\subsubsection{Automatic Evaluation}
\label{app:patient_auto_eval}

\paragraph{Evaluation Setup.}

\begin{table*}[t]
    \centering
    \scriptsize

    \begin{tcolorbox}[
        evalprompt,
        title={Prompt for Automatic Evaluation (Realism Profile)},
        width=0.95\textwidth
    ]
    \ttfamily
    \raggedright

You are a clinical psychiatrist validating a simulated patient agent that is used to test diagnostic-interview systems. Output JSON only, no other text.

\par\smallskip

You are reviewing the symptom profile of a simulated psychiatric patient.

\medskip
\textbf{QUESTION}

Is each symptom manifestation written the way a real patient would actually put it when describing that symptom?

\medskip
\textbf{WHAT YOU SCORE}

Only the quoted sentences under [Symptom Manifestations]. Onset \& Background is there as context for why each symptom is phrased this way-read it, but do not score it. Severity, Additional Criteria, Duration, and Functional Impairment are not scored.

\medskip
\textbf{HOW TO USE [Symptom Definitions]}

[Symptom Definitions] is a JSON object giving, for each numbered manifestation, the symptom it was written from: its clinical \texttt{description}, its \texttt{category}, and its \texttt{subtypes} when the symptom has them.

\par\smallskip

Use it as reference-never score the definitions themselves. Check the manifestation against that information: does the sentence express that symptom, in the form the \texttt{description} and, when present, the matching \texttt{subtype} specify, the way someone actually living it would say it?

\par\smallskip

A sentence that restates the definition in first person, or that describes something adjacent to the symptom rather than the symptom itself, is what you are looking for.

\medskip
\textbf{WHAT TO DO}

First, list by number every manifestation a real patient would not plausibly say that way.

\par\smallskip

Then rate the profile as a whole from 1 to 5:

5 - Every sentence reads like something a real patient would actually say.\\
4 - Every sentence conveys its symptom, but the wording of some is stilted, oddly chosen, or translated-sounding-phrasing a person would not actually use.\\
3 - 1-2 sentences that a real patient would not plausibly say that way.\\
2 - 3 or more such sentences.\\
1 - More than half of the sentences are like that.

A well-written profile is supposed to come back a 5 with an empty list.

\par\smallskip

Finally, give the reason for the rating you gave.

\medskip
[Symptom Definitions - Reference Only, Not Scored]

\{criteria\}

\medskip
[Clinical Profile]

\{profile\}

\medskip
Output JSON only:

\begin{verbatim}
{
  "unrealistic": [
    {
      "n": <number>,
      "name": "<symptom name>",
      "why": "<short>"
    }
  ],
  "rating": <1-5>,
  "reason": "<1-3 sentences in English, why that rating>"
}
\end{verbatim}

    \end{tcolorbox}
    \caption{\label{tab:prompt_automatic_evaluation_realism_profile}Prompt for Automatic Evaluation (Realism Profile)}
\end{table*}
\begin{table*}[t]
    \centering
    \scriptsize

    \begin{tcolorbox}[
        evalprompt,
        title={Prompt for Automatic Evaluation (Realism Conversation)},
        width=0.95\textwidth
    ]
    \ttfamily
    \raggedright

You are a clinical psychiatrist validating a simulated patient agent that is used to test diagnostic-interview systems. Output JSON only, no other text.

\par\smallskip

You are evaluating a psychiatric diagnostic interview conducted by a simulated patient agent.

\medskip
\textbf{QUESTION}

Do the patient agent's utterances feel natural, like those of a real patient?

\medskip
\textbf{WHAT YOU SCORE}

Only the patient agent's utterances. The doctor's questions were also generated by an AI and are not evaluated - read them only to see what the patient is answering.

\par\smallskip

Do not let the doctor repeating a question, or conducting the interview inefficiently, lower the patient's score. When the doctor asks the same thing again, repeating the same answer is what a patient would do - do not weigh it heavily against them.

\medskip
\textbf{SCORING}

(5) The interview reads throughout like the utterances of a real patient.\\
(4) Some expressions are awkward, but overall it reads like a real patient.\\
(3) Parts hard to read as a real patient's utterances appear occasionally.\\
(2) Parts hard to read as a real patient's utterances appear frequently.\\
(1) The interview as a whole is hard to read as a real patient's utterances.

\medskip
\textbf{WHAT TO DO}

Write down the specific utterances that felt awkward, with their turn numbers. Then give the rating, and the reason for it.

\medskip
[Interview Transcript]

\{transcript\}

\medskip
Output JSON only:

\begin{verbatim}
{
  "unnatural": [
    {
      "turn": <n>,
      "why": "<the utterance that felt awkward, and why>"
    }
  ],
  "rating": <1-5>,
  "reason": "<1-3 sentences in English, why that rating>"
}
\end{verbatim}

    \end{tcolorbox}
\caption{\label{tab:prompt_realism_conversation}Prompt for Automatic Evaluation (Realism Conversation) }
\end{table*}
\begin{table*}[t]
    \centering
    \scriptsize

    \begin{tcolorbox}[
        evalprompt,
        title={Prompt for Clinical Profile Fidelity Validation},
        width=0.95\textwidth
    ]
    \ttfamily
    \raggedright

You are a clinical psychiatrist validating a simulated patient agent that is used to test diagnostic-interview systems. Output JSON only, no other text.

\par\smallskip

A simulated patient was given the clinical profile below and told to answer only what the doctor asks, staying within that profile.

\medskip
\textbf{QUESTION}

Does the patient stay faithful to the given clinical profile?

\medskip
\textbf{VIOLATIONS}

(A) Stating clinical information that is not in the profile
(medications, tests, dates, events, other symptoms, ...)\\
(B) Answering in a way that contradicts the profile\\
(C) Confirming a symptom or detail the doctor names that is not in the profile\\
(D) Not presenting the Chief Complaint in the first turn, or replacing it with another symptom

\medskip
\textbf{NOT VIOLATIONS}

\begin{itemize}[
    label=-,
    leftmargin=1.2em,
    labelsep=0.4em,
    itemsep=0pt,
    topsep=0pt
]
    \item Playing a symptom down or understating it
    \item Bringing a symptom that is in the profile up late in the conversation
    \item Saying ``I don't know'' when asked about something the profile does not cover
    \item Some profile symptoms never coming up (the patient is designed not to raise a symptom unless the doctor asks)
\end{itemize}

\par\smallskip

The doctor is also an AI and is not evaluated. Repeated or inefficient questioning must not count against the patient.

\medskip
\textbf{WHAT TO DO}

List each violation with its turn number and type. Then mark whether it bears on the diagnosis — this second judgement is the one that decides the score.

\par\smallskip

`major' - The violation could affect the doctor's diagnostic judgement.\\
`minor' - It does not affect the diagnostic judgement.

\medskip
\textbf{SCORING}

(5) The patient answers faithfully to the given clinical profile.\\
(4) 1-2 minor violations that do not affect the diagnosis.\\
(3) 3 or more minor violations that do not affect the diagnosis.\\
(2) There is a violation that could affect the doctor's diagnostic judgement.\\
(1) The conversation as a whole proceeds unrelated to the profile.

\par\smallskip

Then give the reason for the rating.

\medskip
[Clinical Profile]

A symptom counts as `in the profile' only if it appears here.

\par\smallskip

\{profile\}

\medskip
[Interview Transcript]

\{transcript\}

\medskip
Output JSON only:

\begin{verbatim}
{
  "violations": [
    {
      "turn": <n>,
      "type": "A|B|C|D",
      "severity": "major|minor",
      "note": "<what was said>"
    }
  ],
  "rating": <1-5>,
  "reason": "<1-3 sentences in English, why that rating>"
}
\end{verbatim}

    \end{tcolorbox}

\caption{\label{tab:prompt_automatic_evaluation_fidelity}Prompt for clinical profile fidelity validation}
\end{table*}

To extend the expert evaluation beyond the 46 sampled conversations, we conducted an automated evaluation with an LLM judge on the same three dimensions. We used Claude Sonnet 5 as the judge to reduce self-preference, as it belongs to a model family different from the patient backbones (GPT and Gemini). The judge was given the clinical profile and interview transcript in the same format as the experts, along with the same scoring criteria. For Profile Fidelity, we additionally provided the clinical definition and subtypes of each symptom as reference information in the prompt, explicitly supplying the domain-specific knowledge that clinical experts are assumed to possess so that the judge's assessment is grounded in clinical knowledge. Conversation Realism was assessed on at most the first 15 turns of each conversation (the average conversation length is 11.46 turns), as longer conversations often involve repeated questions from the doctor, which induce repetitive patient responses and lower the score regardless of the simulator's own quality. The prompts are presented in Tables~\ref{tab:prompt_automatic_evaluation_realism_profile}--\ref{tab:prompt_automatic_evaluation_fidelity}.

\paragraph{Evaluation Results.}
For comparability with the expert evaluation, Conversation Realism and Profile Fidelity were assessed on the 2,068 conversations conducted with the same three doctor models as in the expert sample (Claude Sonnet 5, Gemini 3.8 Flash, and GPT 5.4), while Profile Realism was assessed on all 345 profiles independently of the conversations. The simulator achieved high mean ratings of 4.98, 4.78, and 4.81 on Profile Realism, Conversation Realism, and Profile Fidelity, respectively. On the 46 expert-rated cases, the automated ratings also showed high agreement with the majority-voted expert ratings across all three dimensions (AC$_2$ $=$ .97 -- .99; Table~\ref{tab:expert_validation}). This indicates that the LLM judge serves as a reliable proxy for expert judgment, supporting that the high quality observed in the expert sample generalizes to the full set of conversations. Table~\ref{tab:patient_auto_backbone} reports the results by patient backbone. Gemini 3.1 Pro achieved higher Conversation Realism but lower Profile Fidelity than GPT 5.6 Terra, revealing a trade-off between realism and fidelity across backbones.

\begin{table}[H]
    \centering
    \resizebox{\columnwidth}{!}{%
    \begin{tabular}{@{}lccc@{}}
    \toprule
        \textbf{Dimension} & \textbf{Gemini 3.1 Pro} & \textbf{GPT 5.6 Terra} & \textbf{Overall} \\
    \midrule
        Profile Realism      & --   & --   & 4.98 \\
        Conversation Realism & 4.89 & 4.67 & 4.78 \\
        Profile Fidelity     & 4.73 & 4.89 & 4.81 \\
    \bottomrule
    \end{tabular}%
    }
    \caption{Automated evaluation results by patient backbone (mean ratings on a 1--5 scale). Profile Realism is rated per profile and is therefore independent of the patient backbone.}
    \label{tab:patient_auto_backbone}
\end{table}

\subsubsection{Symptom Expression Diversity}
\label{app:patient_symptom_diversity}

\paragraph{Evaluation Setup.}
LLM-generated text tends to be homogeneous in expression~\citep{padmakumar2024does}. If this generator bias appears in our simulator, patients with the same disorder would describe their symptoms in similar ways, and a doctor model could reach the correct diagnosis from surface-level expressions alone. To test this, we measure the diversity of patient utterances with n-gram-based metrics (Distinct-2/3, 1$-$Self-BLEU, and N-gram Vendi) and an embedding-based metric (Embedding Vendi, using all-MiniLM-L6-v2), and compare two conditions: patients with the same disorder (\textit{within disease}) and patients with different disorders (\textit{mixed disease}). Within-disease sets consist of all 15 patients of a disorder (five profiles $\times$ three personas), and mixed-disease sets consist of one patient from each of 15 randomly selected disorders, averaged over 100 draws. Since diversity metrics are sensitive to sample size, both conditions are matched in the numbers of patients (15), conversations (150), and utterances (500). Some reduction in within-disease diversity is natural, as patients with the same disorder share symptoms; a large reduction, however, would indicate that their expressions have collapsed into templates.

\paragraph{Evaluation Results.}
As shown in Table~\ref{tab:diversity}, within-disease diversity retains 78--93\% of mixed-disease diversity across all metrics (e.g., Distinct-2: 0.357 vs.\ 0.404), showing that patients with the same disorder still describe their symptoms in distinct ways. This result suggests that, through context-grounded symptom manifestation, even patients sharing the same target disorder express their symptoms according to their individual backgrounds, rather than following template-like expressions of a specific LLM. Moreover, the two patient backbones show comparable levels of within-disease diversity, indicating that this result does not depend on a particular generator.

\begin{table}[H]
    \centering
    \resizebox{\columnwidth}{!}{%
    \begin{tabular}{@{}lccccc@{}}
    \toprule
        \textbf{Condition} & \textbf{Distinct-2} & \textbf{Distinct-3} & \textbf{1$-$Self-BLEU} & \textbf{Vendi}$_{\text{N-gram}}$ & \textbf{Vendi}$_{\text{Embedding}}$ \\
    \midrule
        Mixed disease
        & 0.404$_{\pm0.008}$ & 0.676$_{\pm0.011}$ & 0.454$_{\pm0.012}$ & 319.1$_{\pm4.6}$ & 51.1$_{\pm1.8}$ \\
    \addlinespace
        Within disease
        & 0.357$_{\pm0.040}$ & 0.619$_{\pm0.053}$ & 0.397$_{\pm0.050}$ & 297.5$_{\pm20.3}$ & 39.8$_{\pm7.6}$ \\
        \quad GPT 5.6 Terra
        & 0.336$_{\pm0.043}$ & 0.562$_{\pm0.053}$ & 0.338$_{\pm0.047}$ & 264.9$_{\pm22.6}$ & 40.6$_{\pm8.2}$ \\
        \quad Gemini 3.1 Pro
        & 0.327$_{\pm0.033}$ & 0.588$_{\pm0.047}$ & 0.374$_{\pm0.043}$ & 286.5$_{\pm18.4}$ & 36.9$_{\pm6.3}$ \\
    \bottomrule
    \end{tabular}%
    }
    \caption{Diversity of patient utterances among patients with the same disorder (\textit{Within disease}) and with different disorders (\textit{Mixed disease}). Indented rows restrict the within-disease condition to a single patient backbone. Both conditions are matched in the numbers of patients, conversations, and utterances. Higher values indicate greater diversity for all metrics.}
    \label{tab:diversity}
\end{table}

\subsection{Case Study}
\label{app:patient_case}

\begin{figure*}[p]
    \centering
    \includegraphics[width=\textwidth]{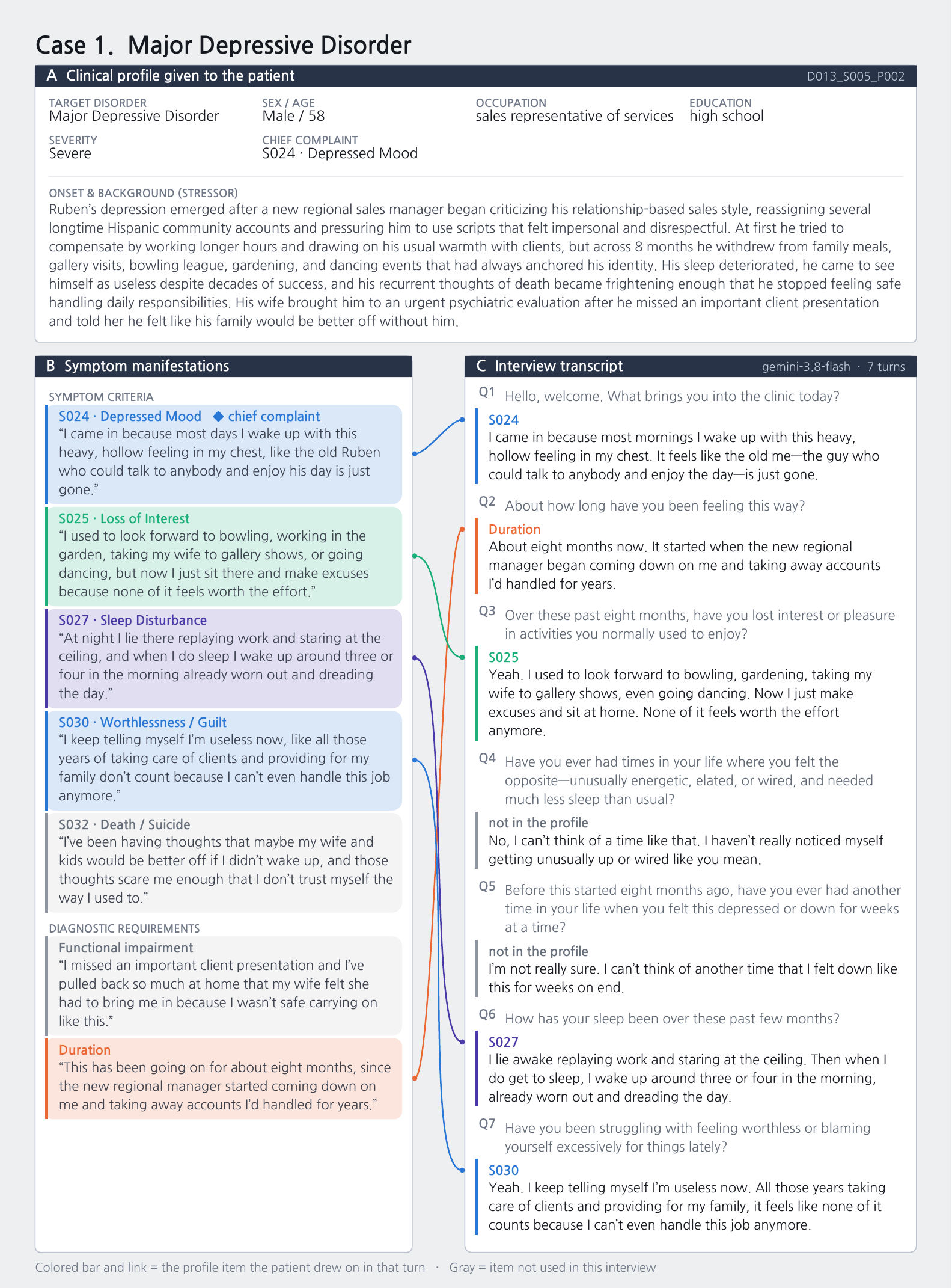}
    \caption{Case study of a simulated patient with major depressive disorder (doctor: Gemini 3.8 Flash). Colored bars and links indicate the profile item the patient drew on in each turn; gray items were not used in the interview. The patient presents the chief complaint in the first turn, discloses only the items targeted by each question, and expresses uncertainty when asked about information not specified in the profile (Q4, Q5).}
    \label{fig:case_mdd}
\end{figure*}

\begin{figure*}[p]
    \centering
    \includegraphics[width=\textwidth,height=0.88\textheight,keepaspectratio]{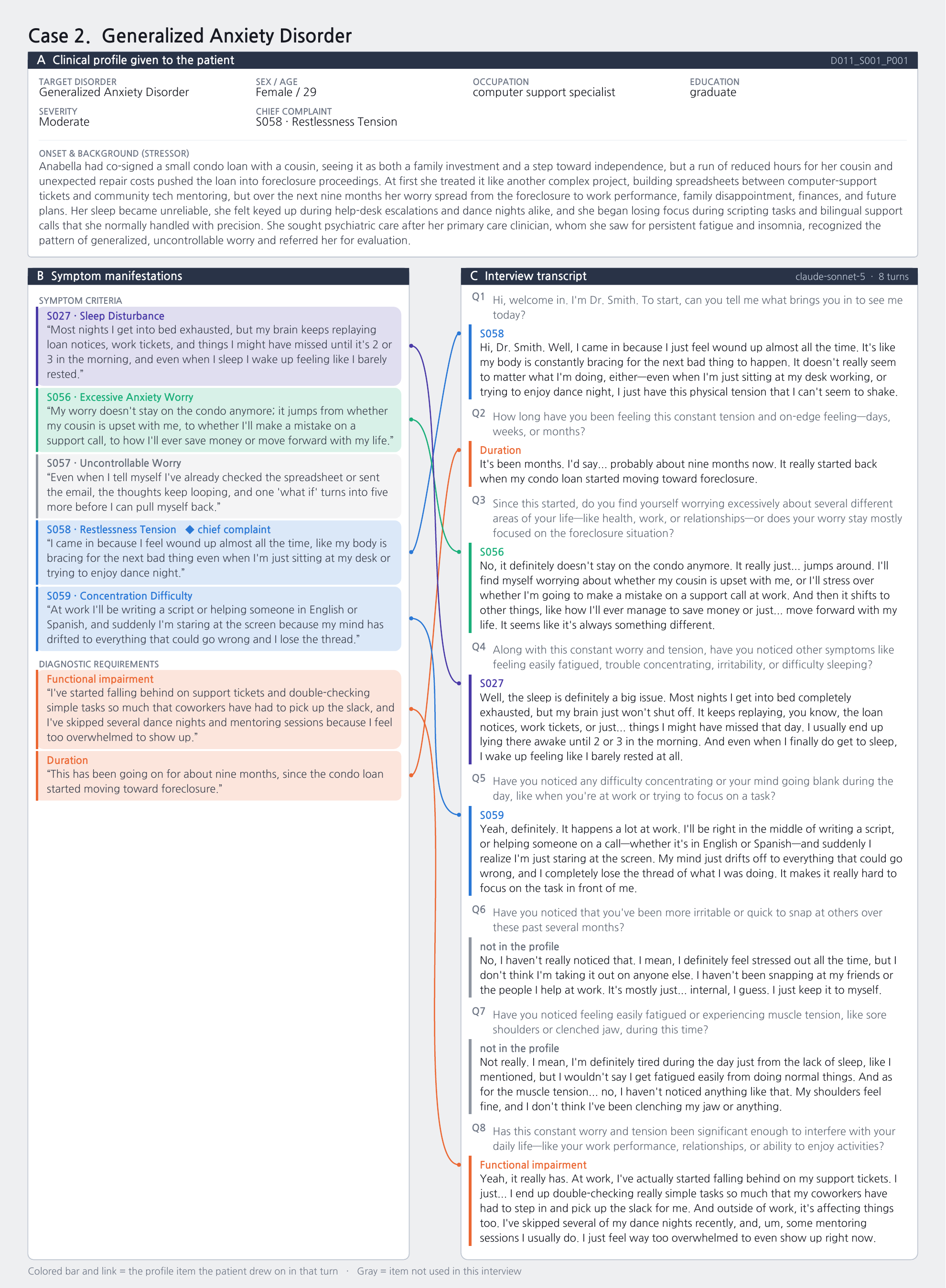}
    \caption{Case study of a simulated patient with generalized anxiety disorder (doctor: Claude Sonnet 5). Colored bars and links indicate the profile item the patient drew on in each turn; gray items were not used in the interview.}
    \label{fig:case_gad}
\end{figure*}

We present two example interviews generated by our simulator. Each figure consists of the clinical profile given to the patient (A), the pre-generated symptom manifestations (B), and the interview transcript (C), where links indicate which profile item the patient drew on in response to each of the doctor's questions. Figures~\ref{fig:case_mdd} and~\ref{fig:case_gad} show actual interviews with patients with major depressive disorder and generalized anxiety disorder, respectively. In both cases, the patient agent refers only to the list of symptom manifestations rather than the clinical profile itself, maps each of the doctor's questions to the corresponding item, and responds based on that item's manifestation. When asked about information not specified in the profile, the patient responds without introducing new diagnostic information, either expressing uncertainty or stating that it has not experienced such symptoms.

\clearpage
\section{Implementation for Doctor LLM}
\label{app:framework_doctor}

We implement the doctor LLM using separate prompts for diagnostic hypothesis formation, follow-up questioning, and final diagnosis. The doctor receives only the dialogue history and its own previous outputs, without access to the hidden clinical profile, target disorder, MentalKG, DSM-5 criteria, or a predefined list of candidate disorders. All outputs follow structured JSON formats to support consistent interaction and evaluation.

\paragraph{Diagnostic Hypothesis Formation}
At each turn $t$, the doctor receives the full dialogue history, its candidate set from the previous turn, and the current and maximum turn numbers. It generates an updated candidate set $\hat{C}_t$ using ICD-10 codes~\cite{world2004icd}, along with a brief note describing changes from the previous set. The doctor also outputs a Boolean indicator, \texttt{is\_final}, specifying whether it is ready to make a final diagnosis. This allows the doctor to update its hypotheses and independently determine when to terminate questioning.
\begin{table}[ht!]
    \centering
    \scriptsize

    \begin{tcolorbox}[
        doctorprompt,
        title={Prompt for Candidate Disease Inference},
        width=0.95\linewidth
    ]
    \ttfamily
    \raggedright

You are a board-certified psychiatrist interviewing a patient to reach a diagnosis.

\par\smallskip

Your goal is to reach the correct diagnosis within \{max\_turns\} turns of talking with the patient. From the dialogue so far and your candidates from the previous turn, list the disorders this patient could have.

\medskip
\textbf{[Current Turn]}

\{current\_turn\}

\medskip
\textbf{[Previous Candidates]}

\{previous\_candidates\}

\medskip
\textbf{[Dialogue History]}

\{dialogue\_history\}

\medskip
\textbf{[Instruction]}

\begin{itemize}[
    label=-,
    leftmargin=1.2em,
    labelsep=0.4em,
    itemsep=0pt,
    topsep=0pt
]
    \item Give candidates as ICD-10 codes, not disease names.
    \item If the list changed from the previous candidates, say in ``note'' what this turn added or ruled out. If it did not change, leave ``note'' empty.
    \item If the candidates have narrowed to one and you are confident enough that no further questioning is needed, return true for is\_final.
\end{itemize}

\medskip
\textbf{[Output Format]}

Output JSON only, no other text.

\begin{verbatim}
{
  "candidates": ["<ICD-10 code>", "..."],
  "note": "<why the list changed, or empty>",
  "is_final": <true or false>
}
\end{verbatim}

    \end{tcolorbox}
    \caption{\label{tab:prompt_candidate_disease_inference}Prompt for candidate disease inference}
    
\end{table}

\begin{table}[ht]
    \centering
    \scriptsize

    \begin{tcolorbox}[
        doctorprompt,
        title={Prompt for Question Generation},
        width=0.95\linewidth
    ]
    \ttfamily
    \raggedright

You are a board-certified psychiatrist interviewing a patient to reach a diagnosis.

\par\smallskip

Your goal is to reach the correct diagnosis within \{max\_turns\} turns of talking with the patient. From the dialogue so far and your current candidates, ask the one follow-up question that best separates them.

\medskip
\textbf{[Current Turn]}

\{current\_turn\}

\medskip
\textbf{[Current Candidates]}

\{current\_candidates\}

\medskip
\textbf{[Dialogue History]}

\{dialogue\_history\}

\medskip
\textbf{[Instruction]}

\begin{itemize}[
    label=-,
    leftmargin=1.2em,
    labelsep=0.4em,
    itemsep=0pt,
    topsep=0pt
]
    \item Ask about exactly ONE thing. Never bundle several symptoms, or several attributes of one symptom, into a single question---e.g., asking whether a symptom is present and how long it has lasted at the same time.
    \item If you need to ask about more than one thing, ask the one that tells you the most now and come back to the rest later.
    \item Keep the question clear enough for the patient to understand.
    \item Respond in English.
\end{itemize}

\medskip
\textbf{[Output Format]}

Output JSON only, no other text. Only ``question'' is delivered to the patient.

\begin{verbatim}
{
  "question": "<the question to ask the patient>",
  "intent": "<the information this question seeks>"
}
\end{verbatim}

    \end{tcolorbox}
    \caption{\label{tab:prompt_question_generation}Prompt for question generation}
    
\end{table}
\paragraph{Follow-up Question Generation}
If the doctor continues the interview, a separate questioning prompt receives the updated candidate set $\hat{C}_t$ and the dialogue history. The doctor generates a follow-up question $q_t$ and an accompanying \texttt{intent} describing the information it seeks. The prompt instructs the doctor to ask about a single clinical aspect, prioritizing information that distinguishes among its current diagnostic candidates. At the beginning of the interview, the doctor instead receives a fixed instruction to greet the patient and ask \textit{``What brings you in today?''}

\paragraph{Final Diagnosis and Diagnostic Evidence Checklist}
The interview terminates when the doctor sets \texttt{is\_final} to \texttt{true} or reaches the maximum turn limit. The final diagnosis prompt receives the complete dialogue history and the most recent candidate set. The doctor then outputs a single final diagnosis as an ICD-10 code, two or three alternative diagnoses considered, and a brief rationale. The final response also includes a diagnostic evidence checklist describing the information the doctor considers to support its diagnosis.

The checklist records the doctor's self-reported diagnostic evidence in a structured format. It includes confirmed symptoms organized by diagnostic symptom groups, with the symptoms and their counts reported for each group. Additional fields record symptom duration, functional impairment, traumatic or psychosocial stressors, and other disorder-specific requirements. Duration is reported as free text, while impairment and stressor fields allow Boolean or null values. The checklist is used in the evaluation of Diagnostic Evidence Sufficiency. Symptom-group information is cross-checked against independently extracted patient evidence, whereas non-symptom requirements are evaluated using the doctor's reported checklist values and the corresponding MentalKG requirements.

\begin{table}[ht]
    \centering
    \scriptsize

    \begin{tcolorbox}[
        doctorprompt,
        title={Prompt for Final Diagnosis},
        width=0.95\linewidth
    ]
    \ttfamily
    \raggedright

    \setlength{\leftmargini}{1.2em}
    \renewcommand{\labelitemi}{-}

You are a board-certified psychiatrist. Your goal is to identify the patient's accurate diagnosis through an interview.

\par\smallskip

Based on the full interview transcript and the differential diagnosis candidates, write the final diagnosis report.

\par\smallskip

Both the ``diagnosis'' and ``candidates'' MUST be ICD-10 codes.

\medskip
\textbf{[Field Instructions]}

\begin{itemize}[
    label=-,
    leftmargin=1.2em,
    labelsep=0.4em,
    itemsep=0pt,
    topsep=0pt
]
    \item diagnosis: The single diagnosis code of ICD-10.
    \item candidates: Top 2-3 differential candidates of ICD-10 codes considered until the end.
    \item reason: Brief diagnostic rationale.
    \item diagnostic\_checklist: Structured evidence summary organized by diagnostic criterion type. List only symptom groups that are relevant to your diagnosis, with the symptoms the patient actually confirmed.
\end{itemize}

\medskip
\textbf{[Output Format]}

Output JSON only, no other text.

\begin{verbatim}
{
  "diagnosis": "<ICD-10 code>",
  "candidates": ["<ICD-10 code>", "..."],
  "reason": "<brief diagnostic rationale>",
  "diagnostic_checklist": {
    "symptom_groups": {
      "<symptom group>": {
        "symptoms": ["<confirmed symptom>", "..."],
        "count": <int>
      }
    },
    "duration": "<free text, or null>",
    "functional_impairment": <true | false | null>,
    "traumatic_stressor": <true | false | null>,
    "psychosocial_stressor": <true | false | null>,
    "additional_requirements": ["<requirement met>", 
                                "..."]
  }
}
\end{verbatim}

    \end{tcolorbox}
    \caption{\label{tab:prompt_final_diagnosis}Prompt for final diagnosis}
    
\end{table}



\section{Implementation for Evaluation}
\label{app:evaluation_implementation}

We provide implementation details for the four process-level evaluation dimensions described in Section~\ref{sec:process_evaluation}. The evaluation combines LLM-based extraction of diagnostic information with rule-based scoring grounded in MentalKG. 

\subsection{Evaluation Setup}

\paragraph{Dataset Statistics}
\begin{table}[ht!]
    \centering
    \small
    \resizebox{\linewidth}{!}{
    \begin{tabular}{@{}lrr@{}}
    \toprule
        \textbf{Statistic} & \multicolumn{2}{r}{\textbf{Value}} \\
    \midrule
        \multicolumn{3}{@{}l}{\textbf{\textit{Clinical profiles and patients}}} \\
        Disorders & \multicolumn{2}{r}{23} \\
        Clinical profiles (5 per disorder) & \multicolumn{2}{r}{115} \\
        Patients (3 per profile) & \multicolumn{2}{r}{345} \\
        Symptoms per profile & \multicolumn{2}{r}{6.1 (1--15)} \\
        Additional requirements per profile & \multicolumn{2}{r}{0.7 (0--3)} \\
        Profiles requiring functional impairment & \multicolumn{2}{r}{65} \\
        Sex (female / male) & \multicolumn{2}{r}{173 / 172} \\
        Age & \multicolumn{2}{r}{45.2 (18--87)} \\
        Severity (mild / moderate / severe) & \multicolumn{2}{r}{115 / 115 / 115} \\
        Distinct life events & \multicolumn{2}{r}{31} \\
        Distinct chief-complaint symptoms & \multicolumn{2}{r}{56} \\
        Words per symptom manifestation & \multicolumn{2}{r}{35.1 (21--67)} \\
    \addlinespace
    \midrule
    \end{tabular}
    }
    \resizebox{\linewidth}{!}{
    \begin{tabular}{@{}lrr@{}}
        \multicolumn{3}{@{}l}{\textbf{\textit{Interview conversations}}} \\
        & \textbf{GPT 5.6 Terra} & \textbf{Gemini 3.1 Pro} \\
        Doctor models & 5 & 5 \\
        Conversations & 1,723 & 1,724 \\
        Patient turns per conversation & 12.3 (2--20) & 10.6 (2--20) \\
        Words per patient utterance & 40.1 & 60.0 \\
        Words per doctor utterance & 18.9 & 19.2 \\
    \bottomrule
    \end{tabular}
    }
    %
    \caption{Statistics of the clinical profiles, simulated patients, and interview conversations. Values are means with ranges in parentheses unless noted. Conversation statistics are computed over all conversations.}
    \label{tab:data_statistics}
\end{table}

\autoref{tab:data_statistics} summarizes the resulting clinical profiles, patients, and interview conversations.

\paragraph{Inference configuration}
Closed-source models are accessed through API services
\footnote{
GPTs: \url{https://platform.openai.com/}\\
Geminis: \url{https://ai.google.dev/api}\\
Claudes: \url{https://openrouter.ai/}
}, whereas open-source models are deployed locally using vLLM\footnote{\url{https://vllm.ai/}} on a server equipped with 8 NVIDIA H100 GPUs. For open-source models, we follow their default generation configurations without additional decoding hyperparameter tuning.

\subsection{Implementation and Calculation of Metrics}

\paragraph{Turn-Level Symptom Evidence Extraction}
\begin{table*}[t]
    \centering
    \resizebox{\textwidth}{!}{
    \scriptsize
    \begin{tabular}{l p{5.5cm} l p{7.5cm}}
    \toprule
    ID & Disorder (List from MentalKG; \citet{song2026mentalbench}) & ICD-10 & Accepted ICD-10 codes \\
    \midrule
D001 & Attention-Deficit/Hyperactivity Disorder (Combined Presentation) & F90.2 & F90.2, F90.8, F90.9 \\
D002 & Attention-Deficit/Hyperactivity Disorder (Predominantly Inattentive Presentation) & F90.0 & F90.0 \\
D003 & Attention-Deficit/Hyperactivity Disorder (Predominantly Hyperactive/Impulsive Presentation) & F90.1 & F90.1 \\
D004 & Delusional Disorder & F22 & F22, F22.0, F22.8, F22.9 \\
D005 & Schizophrenia & F20.9 & F20, F20.0, F20.1, F20.2, F20.3, F20.5, F20.6, F20.89, F20.9 \\
D006 & Schizoaffective Disorder (Bipolar Type) & F25.0 & F25, F25.0, F25.9 \\
D007 & Schizoaffective Disorder (Depressive Type) & F25.1 & F25.1 \\
D008 & Bipolar I Disorder & F31.9 & F30.1, F30.9, F31, F31.0, F31.10, F31.11, F31.12, F31.13, F31.3, F31.30, F31.31, F31.32, F31.4, F31.6, F31.60, F31.61, F31.62, F31.63, F31.70, F31.71, F31.72, F31.73, F31.74, F31.75, F31.76, F31.77, F31.78, F31.89, F31.9 \\
D009 & Bipolar II Disorder & F31.81 & F31.8, F31.81 \\
D010 & Bipolar I Disorder with Psychotic Features & F31.2 & F30.2, F31.1, F31.2, F31.5, F31.64 \\
D011 & Generalized Anxiety Disorder & F41.1 & F41.1 \\
D012 & Specific Phobia & F40.9 & F40.0, F40.00, F40.01, F40.1, F40.10, F40.2, F40.20, F40.201, F40.21, F40.210, F40.218, F40.22, F40.228, F40.230, F40.231, F40.232, F40.233, F40.241, F40.242, F40.243, F40.248, F40.29, F40.290, F40.298, F40.9 \\
D013 & Major Depressive Disorder & F32.9 & F32, F32.0, F32.1, F32.2, F32.4, F32.5, F32.9, F32.A, F33, F33.0, F33.1, F33.2, F33.40, F33.41, F33.42, F33.9 \\
D014 & Persistent Depressive Disorder & F34.1 & F34.1 \\
D015 & Major Depressive Disorder with Psychotic Features & F32.3 & F32.3, F33.3 \\
D016 & Obsessive-Compulsive Disorder & F42.2 & F42, F42.0, F42.1, F42.2, F42.8, F42.9 \\
D017 & Body Dysmorphic Disorder & F45.22 & F45.2, F45.22 \\
D018 & Posttraumatic Stress Disorder & F43.10 & F43.1, F43.10, F43.11, F43.12 \\
D019 & Acute Stress Disorder & F43.0 & F43.0, F43.8 \\
D020 & Adjustment Disorder & F43.20 & F43.2, F43.20, F43.21, F43.22, F43.23, F43.24, F43.25, F43.29 \\
D021 & Anorexia Nervosa & F50.00 & F50.0, F50.00, F50.01, F50.02, F50.1 \\
D022 & Bulimia Nervosa & F50.2 & F50.2, F50.20, F50.3 \\
D023 & Binge-Eating Disorder & F50.81 & F50.8, F50.81 \\
    \bottomrule
    \end{tabular}
    }
    \caption{\label{tab:icd10_codes}ICD-10 codes of the 23 MentalKG disorders. \textit{ICD-10} is the representative code of each disorder. Every ICD-10 code the doctor outputs, both in its candidate sets and in its final diagnosis, is mapped to the disorder whose \textit{accepted ICD-10 codes} contain it. Each code is accepted for at most one disorder, so subtype codes that MentalKG does not distinguish (e.g., the F20.x subtypes of schizophrenia) are mapped to their parent disorder, whereas codes naming a different diagnosis (e.g., F42.3 hoarding disorder) are excluded.}
\end{table*}

Each interview begins with a fixed opening question (\textit{``What brings you in today?''}) that is excluded from the evaluation. At turn $t \geq 1$, the patient provides response $r_t$, after which the doctor predicts its candidate set $\hat{C}_t$ and, unless the interview terminates, asks a follow-up question $q_t$. Thus, an interview with $T$ patient responses contains $T$ candidate predictions and at most $T-1$ evaluated follow-up questions.

An LLM judge in \ours{} extracts diagnostic symptoms from each patient response and classifies them as confirmed or denied (\autoref{tab:prompt_judge_symptom_extraction}). Symptoms not identified in either category remain unknown. These labels are accumulated across turns to construct the evidence state $E_t$, which is used for reference candidate construction and question evaluation. The doctor reports diagnoses using ICD-10 codes~\cite{world2004icd}, which are mapped to the corresponding disorder identifiers (\autoref{tab:disease_list}) using a predefined mapping that includes accepted alternative codes (\autoref{tab:icd10_codes}). Codes that do not map to any of the 23 disorders are discarded from the candidate sets. An unmapped final diagnosis is counted as incorrect and has no DES; we report how often this occurs in Subsection~\ref{app:detailed_results}.

\begin{table*}[t]
    \centering
    \scriptsize

    \begin{tcolorbox}[
        evalprompt,
        title={Prompt for Symptom Extraction (Evaluation Judge)},
        width=0.95\textwidth
    ]
    \ttfamily
    \raggedright

\textbf{[System]}

You are a clinical psychiatrist. Output only valid JSON.

\medskip
\textbf{[User]}

You are a clinical psychiatrist reviewing one exchange from a patient interview.

\medskip
\textbf{=== SYMPTOM CATALOGUE ===}

\{symptom\_catalogue\}

\medskip
\textbf{=== DOCTOR'S QUESTION (this response is answering) ===}

\{doctor\_question\}

\medskip
\textbf{=== SYMPTOM THIS RESPONSE IS TARGETING ===}

(from the patient's own alignment step --- a hint for grounding, not a fact to assert on its own)

\par\smallskip

\{aligned\_variable\}

\medskip
\textbf{=== PATIENT'S RESPONSE ===}

\{patient\_response\}

\medskip
\textbf{Task:}

\begin{itemize}[
    label=-,
    leftmargin=1.2em,
    labelsep=0.4em,
    itemsep=0pt,
    topsep=0pt
]
    \item Use the doctor's question and the targeted symptom (if given) to correctly read short or ambiguous responses (e.g., ``yeah'', ``not really'', ``sometimes'') --- they are answering THIS question about THIS symptom, not a different one.
    \item CONFIRMED: symptoms clearly present or strongly implied by the patient's words.
    \item DENIED: symptoms the patient explicitly says they do NOT have (e.g., ``No'', ``I don't'', ``I haven't'', ``never'').
    \item Do NOT speculate beyond what the response actually supports. Err toward UNKNOWN (omit) rather than guessing --- this applies to the targeted symptom too: only confirm/deny it if the response itself actually does so, not just because it was the question's target.
\end{itemize}

\medskip
\textbf{[Output Format]}

Reply ONLY with valid JSON (no markdown, no extra text):

\begin{verbatim}
{
  "confirmed": ["S001", ...],
  "denied": ["S025", ...],
  "reasoning": {
    "S001": "one-sentence justification",
    ...
  }
}
\end{verbatim}

    \end{tcolorbox}
    \caption{\label{tab:prompt_judge_symptom_extraction}Prompt for per-turn symptom extraction by the evaluation judge.
    \{symptom\_catalogue\} lists all MentalKG symptoms with their IDs, names, descriptions, and subtypes.
    \{aligned\_variable\} is the profile item selected in the patient simulator's alignment stage, together with its manifestation; it serves only as a hint for interpreting short or ambiguous replies and is set to \textit{none} when no item is aligned.
    The confirmed and denied symptom IDs are accumulated into $\mathrm{Conf}_t$ and $\mathrm{Den}_t$ (Appendix~\ref{app:reference_candidates}).}

\end{table*}

\paragraph{Reference Candidate Set Construction}
\label{app:reference_candidates}

We construct the reference candidate set $C_t$ from the diagnostic evidence $E_t$ accumulated through turn $t$. MentalKG~\cite{song2026mentalbench} specifies mandatory and optional symptom groups for each disorder, including minimum symptom counts and other group-specific requirements. Based on confirmed and denied symptoms reported by a patient, we assign disorders to four categories.

\begin{itemize}
    \item \textit{Excluded}: A mandatory symptom is explicitly denied. Disorders with explicitly denied mandatory symptoms are excluded before assignment to the remaining likelihood categories.
    \item \textit{High likelihood}: All mandatory symptom groups satisfy their minimum symptom counts.
    \item \textit{Moderate likelihood}: At least one mandatory symptom is confirmed, but not all mandatory groups are satisfied.
    \item \textit{Low likelihood}: At least one optional symptom is confirmed, without confirmation of a mandatory symptom.
\end{itemize}

Disorders without relevant symptom evidence are not assigned to a candidate category. Let $H_t$, $M_t$, and $L_t$ denote the sets of disorders assigned to high, moderate, and low likelihood, respectively. We construct the reference candidate set by selecting the highest non-empty category and including the target disorder $d^*$:

\[
C_t =
\begin{cases}
H_t \cup \{d^*\}, & H_t \neq \emptyset,\\
M_t \cup \{d^*\}, & M_t \neq \emptyset,\\
L_t \cup \{d^*\}, & L_t \neq \emptyset,\\
\{d^*\}, & \text{otherwise}.
\end{cases}
\]

Including the target disorder ensures that it remains in the reference candidate set even when the tier-based construction does not select it. The reference set is updated after each patient response using the accumulated evidence, rather than the patient's full hidden profile.

\subsubsection{\dhq{Diagnostic Hypothesis Quality}}
At each turn, we compare the doctor's predicted candidate set $\hat{C}_t$ with the reference set $C_t$. We compute Jaccard, Precision, and Recall as
\[
\mathrm{Jaccard}_t =
\frac{|\hat{C}_t \cap C_t|}
{|\hat{C}_t \cup C_t|},
\]
\[
\mathrm{Precision}_t =
\frac{|\hat{C}_t \cap C_t|}
{|\hat{C}_t|},
\]
\[
\mathrm{Recall}_t =
\frac{|\hat{C}_t \cap C_t|}
{|C_t|}.
\]

Precision is defined as zero when the predicted candidate set is empty. These metrics capture overall candidate agreement, unsupported inclusions, and omissions of reference candidates. For each metric, we first average turn-level scores within each interview and then average the interview-level scores across all interviews.

\subsubsection{\dqq{Question Targeting Score}}
\label{app:question_targeting_score}

\begin{table}[t]
    \centering
    \scriptsize

    \begin{tcolorbox}[
        evalprompt,
        title={Prompt for Question Target Identification (Evaluation Judge)},
        width=0.95\linewidth
    ]
    \ttfamily
    \raggedright

\textbf{[System]}

You are a clinical symptom analyst. Given a doctor's question and a list of candidate psychiatric symptoms, identify which symptom IDs the question is directly attempting to assess. Output ONLY valid JSON --- no explanation, no markdown.

\medskip
\textbf{[User]}

\textbf{DOCTOR'S QUESTION:}

\{doctor\_question\}

\medskip
\textbf{CANDIDATE SYMPTOMS (top \{k\} by semantic similarity):}

\{candidate\_symptoms\}

\medskip
\textbf{[Output Format]}

Reply ONLY with JSON:

\begin{verbatim}
{"symptom_ids": ["S001", ...]}
\end{verbatim}

Include only IDs from the candidate list above that the question clearly targets.

    \end{tcolorbox}
    \caption{\label{tab:prompt_judge_question_targeting}Prompt for identifying the symptoms targeted by the doctor's question. The candidate symptoms are pre-selected from the full symptom catalog by embedding similarity to the question.}

\end{table}

\paragraph{Question Targeting Score} (QTS) evaluates whether the doctor's follow-up question targets diagnostically relevant and unresolved symptoms. At turn $t$, we first map the question $q_t$ to a set of targeted symptom identifiers $S(q_t)$. To identify these symptoms efficiently, we use cosine similarity to select the top-$K$ most relevant symptoms from the 84 symptoms in MentalKG ($K=30$ in our experiments). An LLM judge then selects the symptoms actually targeted by the question from this subset (\autoref{tab:prompt_judge_question_targeting}). The mapping results are cached by question text to avoid repeated LLM calls.

QTS is computed using the doctor's own candidate set $\hat{C}_t$ and the accumulated symptom evidence $E_t$, rather than the reference candidate set $C_t$. This design allows question quality to be evaluated separately from hypothesis correctness. Even if the doctor predicts incorrect diagnostic candidates, its questions are rewarded when they appropriately target information relevant to its own hypotheses. By evaluating questions relative to $\hat{C}_t$, QTS distinguishes the ability to formulate diagnostically useful questions from the ability to identify plausible diagnostic candidates.

We identify two types of diagnostically relevant symptoms from $\hat{C}_t$. First, discriminative symptoms are associated with some, but not all, of the doctor's candidate disorders:
\begin{align*}
& S_{\mathrm{disc}}(\hat{C}_t) = \\
& \left\{
s \,\middle|\,
0 <
\left|
\left\{
d \in \hat{C}_t : s \in \mathrm{pool}(d)
\right\}
\right|
< |\hat{C}_t|
\right\},
\end{align*}
where $\mathrm{pool}(d)$ includes the mandatory and optional symptoms associated with disorder $d$. This set is empty when fewer than two candidate disorders are reported.

Second, required symptoms are the mandatory symptoms associated with any of the doctor's candidate disorders:
\[
S_{\mathrm{req}}(\hat{C}_t)
=
\bigcup_{d \in \hat{C}_t}
\mathrm{must\_include}(d).
\]
Required symptoms are identified individually, without applying the minimum-count conditions of their symptom groups. 
The diagnostically relevant symptom set is then
\[
S_{\mathrm{rel}}(\hat{C}_t) = S_{\mathrm{disc}}(\hat{C}_t) \cup S_{\mathrm{req}}(\hat{C}_t).
\]

We calculate the Diagnostic Relevance Score (DRS) as the proportion of targeted symptoms belonging to $S_{\mathrm{rel}}(\hat{C}_t)$:
\[
\mathrm{DRS}_t =
\frac{|S(q_t) \cap S_{\mathrm{rel}}(\hat{C}_t)|}{|S(q_t)|}.
\]

We additionally calculate a Redundancy Penalty (RP) to account for questions targeting symptoms already resolved during the interview. Let $E_t^+$ and $E_t^-$ denote the cumulative confirmed and denied symptoms independently extracted from patient responses by the evaluation judge:
\[
\mathrm{RP}_t =
\frac{
|S(q_t) \cap (E_t^+ \cup E_t^-)|
}{
|S(q_t)|
}.
\]
Symptoms that remain unknown are not penalized, allowing the doctor to clarify ambiguous patient responses.

The final Question Targeting Score combines diagnostic relevance and non-redundancy:
\[
\mathrm{QTS}_t =
\mathrm{DRS}_t(1-\mathrm{RP}_t).
\]
QTS is defined as zero when $S(q_t)$ is empty. The score ranges from zero to one, rewarding questions that target diagnostically relevant symptoms while avoiding information already confirmed or denied. Questions targeting multiple symptoms are evaluated based on the proportions of relevant and redundant targets.

QTS is computed for each follow-up question, excluding the fixed opening question and the final turn, which has no subsequent question. We first average turn-level QTS within each interview and then average the interview-level scores across all interviews.

\begin{table*}[t]
    \centering
    \scriptsize

    \begin{tcolorbox}[
        evalprompt,
        title={Prompt for Non-Symptom Requirement Verification (Evaluation Judge)},
        width=0.95\textwidth
    ]
    \ttfamily
    \raggedright

\textbf{[System]}

You are a clinical evaluation judge. Your task is to assess whether a doctor's diagnostic checklist covers specific NON-SYMPTOM diagnostic requirements for a psychiatric disorder: duration, functional impairment, traumatic/psychosocial stressor, and any additional requirements.

\par\smallskip

Symptom-group coverage (mandatory/optional symptom pools) is scored algorithmically from the interview's actual confirmed/denied evidence --- do NOT evaluate symptoms here, only the non-symptom fields below.

\par\smallskip

Output ONLY valid JSON, exactly in the format specified. No preamble or extra text.

\medskip
\textbf{[User]}

\textbf{=== GROUND TRUTH NON-SYMPTOM REQUIREMENTS ===}

\{required\_criteria\}

\medskip
\textbf{=== DOCTOR'S DIAGNOSTIC CHECKLIST (non-symptom fields) ===}

\{doctor\_checklist\}

\medskip
\textbf{=== EVALUATION TASK ===}

Evaluate only the non-symptom fields above (Duration, Functional Impairment, Traumatic/Psychosocial Stressor, Additional Requirements). If a requirement is not listed under GROUND TRUTH, its field MUST be null.

\medskip
\textbf{[Output Format]}

Return this JSON (no other text):

\begin{verbatim}
{
  "duration_verified": <true | false | null (null if no duration
      requirement)>,
  "functional_impairment_verified": <true | false | null (null if not
      required by criteria)>,
  "traumatic_stressor_verified": <true | false | null (null if not
      required)>,
  "psychosocial_stressor_verified": <true | false | null (null if not
      required)>,
  "additional_requirements_coverage": <float 0.0-1.0 | null (null if no
      additional requirements)>,
  "additional_requirements_notes": "<brief note on what was / was not
      covered, or null>"
}
\end{verbatim}

    \end{tcolorbox}
    \caption{\label{tab:prompt_judge_diagnostic_checklist}Prompt for verifying the non-symptom requirements in the doctor's diagnostic checklist.
    The ``ground truth'' requirements are the MentalKG criteria of the doctor's predicted disorder $d^*$, not of the target disorder.}

\end{table*}

\subsubsection{\ddq{Diagnostic Decision Quality}}
\label{app:decision_quality}
\paragraph{Final Accuracy}
assesses whether the doctor's final diagnosis $\hat{d}$ matches the target disorder $d^*$. The final diagnosis, reported as an ICD-10 code, is mapped to its corresponding MentalKG disorder identifier. Each interview receives a binary score, and Final Accuracy is reported as the percentage of correctly diagnosed interviews.

\paragraph{Diagnostic Evidence Sufficiency} (DES)
measures how completely the evidence acquired during the interview satisfies the requirements of the doctor's predicted disorder $\hat{d}$, regardless of whether it matches the target disorder. DES combines symptom-criteria satisfaction with the fulfillment of non-symptom diagnostic requirements.

For symptom criteria, we use the set of confirmed symptoms accumulated by the final turn, $E_T^+$. Based on the diagnostic requirements of the predicted disorder $\hat{d}$ in MentalKG, we identify mandatory and optional symptoms. Let $S_{\mathrm{mand}}(\hat{d})$ and $S_{\mathrm{opt}}(\hat{d})$ denote these symptom sets, respectively. A symptom is considered satisfied if it is included in $E_T^+$.

We calculate the Symptom Satisfaction Score (SymSat) as
\[
\mathrm{SymSat} =
\frac{
2|S_{\mathrm{mand}}(\hat{d}) \cap E_T^+|
+
|S_{\mathrm{opt}}(\hat{d}) \cap E_T^+|
}{
2|S_{\mathrm{mand}}(\hat{d})|
+
|S_{\mathrm{opt}}(\hat{d})|
}.
\]
Mandatory symptoms receive twice the weight of optional symptoms to reflect their greater importance in the diagnostic criteria.

Non-symptom requirements are assessed using an LLM judge. The judge compares the doctor's final diagnostic checklist with the requirements associated with the predicted disorder in MentalKG (\autoref{tab:prompt_judge_diagnostic_checklist}). These requirements may include symptom duration, functional impairment, traumatic or psychosocial stressors, and additional disorder-specific criteria. Each applicable requirement is evaluated using binary scoring, with a score of 1 if the requirement is satisfied and 0 otherwise. The evaluation uses the doctor's reported checklist, or its final rationale when the checklist is unavailable.

DES is computed as the unweighted mean of symptom satisfaction and the binary scores for all applicable non-symptom requirements:

\[
\mathrm{DES} =
\frac{
\mathrm{SymSat}
+
\sum_{j \in \mathcal{R}_{\hat{d}}} s_j
}{
1 + |\mathcal{R}_{\hat{d}}|
},
\]
where $\mathcal{R}_{\hat{d}}$ denotes the set of applicable non-symptom requirements for the predicted disorder $\hat{d}$, and $s_j \in \{0,1\}$ indicates whether requirement $j$ is satisfied according to the LLM judge.

By evaluating the criteria of the predicted disorder, DES measures the extent to which the doctor's own diagnostic decision is supported by the acquired and reported evidence, independently of final diagnostic correctness. Scores are averaged across interviews with valid predicted disorder identifiers.

\begin{table*}[t]
    \centering
    \scriptsize

    \begin{tcolorbox}[
        evalprompt,
        title={Prompt for LLM-based Rating in Metric Validation},
        width=0.95\textwidth
    ]
    \ttfamily
    \raggedright

Score a doctor's psychiatric intake interview on 3 dimensions, 1-5 each.

\medskip
\textbf{[h] Diagnostic Hypothesis Quality} (every turn): does the differential (DDx) at that point sensibly reflect info so far --- no major omissions, no stale irrelevant candidates, updates appropriately on new evidence? NOT about final diagnosis correctness.\\
1=major omissions/stale candidates, no updating.\\
3=reasonable but some gaps or partial updating.\\
5=well-formed at each point, updates cleanly on new evidence.

\medskip
\textbf{[q] Diagnostic Question Quality} (every turn with a Q; T0 is not scored): does the question help rule out or confirm current DDx candidates by targeting their key required symptoms, without repeating already-resolved info? Not about how much info was eventually gathered.\\
1=mostly unhelpful/repetitive.\\
3=somewhat useful but unfocused or some repetition.\\
5=targets the sharpest current uncertainty efficiently, no repetition.

\medskip
\textbf{[evidence\_sufficiency]} (once, whole case): does the evidence actually gathered in the dialogue (see checklist: have= vs missing=) sufficiently support the doctor's final diagnosis per DSM-5 (required symptoms met, duration/functional/additional requirements met)? Mention alone isn't enough --- judge by have= vs missing= counts.\\
1=key evidence largely missing.\\
3=partially supported, notable gaps remain.\\
5=well supported, dialogue alone justifies the diagnosis.

\medskip
Disease legend: \{disease\_legend\}

\medskip
\textbf{--- Dialogue ---}

\{dialogue\}

\medskip
\textbf{--- Final diagnosis vs. DSM-5 checklist ---}

\{evidence\_checklist\}

\medskip
\textbf{[Output Format]}

Reply with ONLY this JSON, no prose:

\begin{verbatim}
{
  "turns": [
    {"t": <int, from 1>, "h": <1-5>,
     "q": <1-5 or null, null if that turn has no Q>},
    ...
  ],
  "evidence_sufficiency": <1-5>
}
\end{verbatim}

    \end{tcolorbox}
    \caption{\label{tab:prompt_judge_metric_validation}Prompt for the LLM-based ratings used in metric validation. Each dialogue turn is given as the patient's answer, the doctor's candidate set, and the doctor's next question. Only the [h] and [q] ratings are analyzed in this paper.}

\end{table*}

\subsection{Expert Validation of Process Metrics}
\label{app:metric_validation}

We examine whether Jaccard and QTS align with expert assessments of diagnostic hypothesis and question quality. Two psychiatrists evaluate 397 diagnostic hypothesis sets and 351 follow-up questions from 46 interview episodes using five-point Likert scales.

\paragraph{Rating Criteria}
The expert rating criteria reflect the evaluation objectives of the corresponding metrics. For \textit{hypothesis quality}, experts assess whether the doctor's candidate set appropriately represents the diagnoses supported by the available evidence. Ratings penalize both excessive inclusion of unsupported diagnoses and omission of plausible diagnoses, matching the two error types captured by Jaccard.

For \textit{question quality}, experts assess whether the follow-up question targets unresolved information that helps distinguish among the doctor's own candidate diagnoses or checks their mandatory symptom requirements. Questions that revisit information already confirmed or ruled out receive lower ratings, consistent with the redundancy penalty in QTS.

\paragraph{Correlation Analysis}
For each evaluated item, we average the ratings of the two psychiatrists and compute Pearson correlations between these mean ratings and the corresponding metric scores. We also use Claude Sonnet 5 as an LLM-based judge under the same evaluation criteria (\autoref{tab:prompt_judge_metric_validation}) and compute correlations between its ratings and our metrics.

\paragraph{Results}
As shown in \autoref{tab:expert_validation_judge}, Jaccard correlates positively with expert-rated hypothesis quality ($r=0.43$), and QTS correlates positively with expert-rated question quality ($r=0.46$; both $p<0.001$). The corresponding correlations with automated ratings are $0.55$ and $0.52$, respectively (both $p<0.001$).

Cross-dimension correlations are substantially weaker. Jaccard correlates with question quality at $r=0.07$ for expert ratings and $r=0.04$ for automated ratings. QTS correlates with hypothesis quality at $r=0.06$ and $r=0.01$, respectively. This pattern supports the distinction between the appropriateness of diagnostic hypotheses and the quality of questions selected given those hypotheses.

\subsection{Case Study}
\label{app:doctor_case_study}

We present three interviews to illustrate how the diagnostic behaviors observed in the main results appear in individual interviews.

\begin{figure*}[p]
    \centering
    \includegraphics[width=\textwidth,height=0.82\textheight,keepaspectratio]{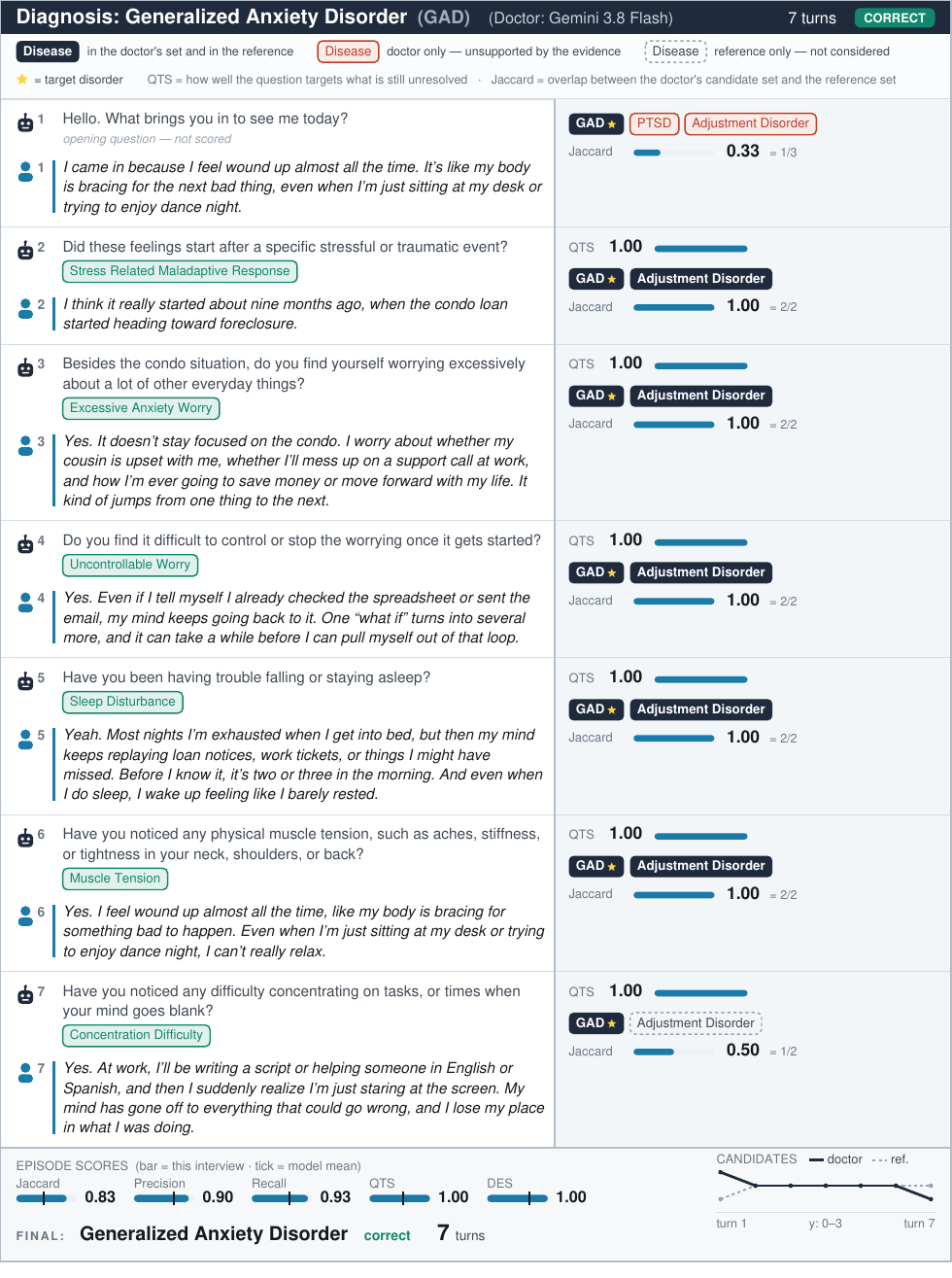}
    \caption{Case study of a diagnostic interview conducted by Gemini 3.8 Flash with a patient with generalized anxiety disorder (GAD).}
    \label{fig:doctor_case_gemini}
\end{figure*}

\paragraph{Case 1: Efficient question targeting.}
In the main results, Gemini 3.8 Flash achieves the highest QTS and the shortest interviews (\autoref{sec:result_by_models}). \autoref{fig:doctor_case_gemini} shows this pattern in an interview with a GAD patient. After the patient discloses a stressful life event (Turn 2), the doctor narrows its candidate set to GAD and adjustment disorder, matching the reference set. Each subsequent question targets an unresolved, diagnostically relevant symptom (QTS $= 1.00$ at every scored turn), and the doctor reaches the correct diagnosis in seven turns. 

\begin{figure*}[p]
    \centering
    \includegraphics[width=\textwidth,height=0.82\textheight,keepaspectratio]{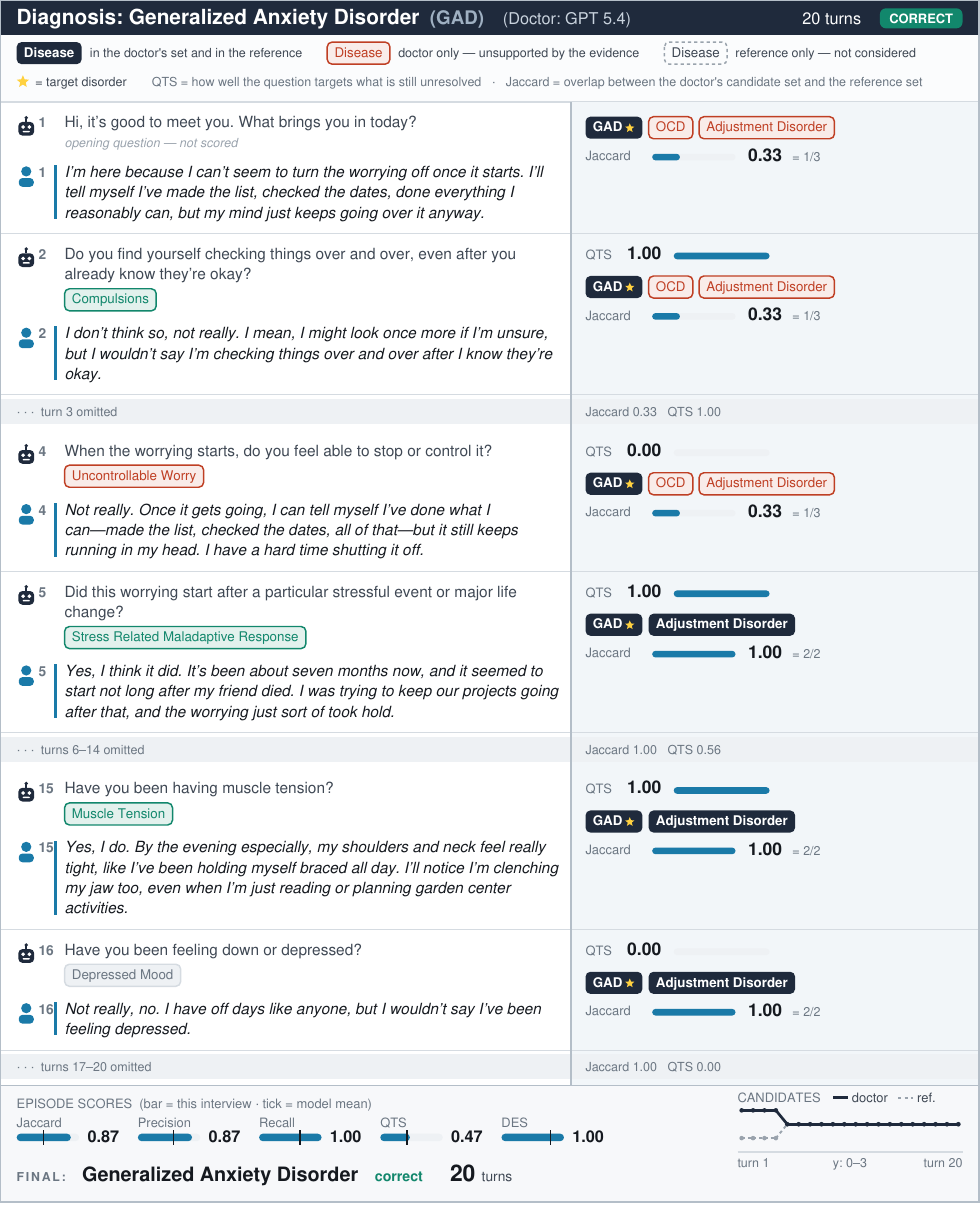}
    \caption{Case study of a diagnostic interview conducted by GPT 5.4 with a patient with generalized anxiety disorder (GAD).}
    \label{fig:doctor_case_gpt}
\end{figure*}

\paragraph{Case 2: Broad coverage and prolonged questioning.}
\autoref{fig:doctor_case_gpt} shows an interview conducted by GPT 5.4 with another GAD patient. The doctor also reaches the correct diagnosis, but its process reflects the broader hypotheses and longer interviews observed in the main results. Early in the interview, it retains OCD, which is not supported by the reference set, and asks again about uncontrollable worry (Turn 4), which the patient had already described in the first turn (QTS $= 0.00$). Although its candidate set matches the reference set from Turn 5 onward, the doctor continues questioning until it uses all 20 turns, and many of these later questions do not help distinguish its current candidates (Turn 16 and Turns 17--20; QTS $= 0.00$). Compared with Case 1, this interview reaches the same correct diagnosis but differs markedly in QTS ($0.47$ vs.\ $1.00$) and interview length (20 vs.\ 7 turns).

\begin{figure*}[p]
    \centering
    \includegraphics[width=\textwidth,height=0.82\textheight,keepaspectratio]{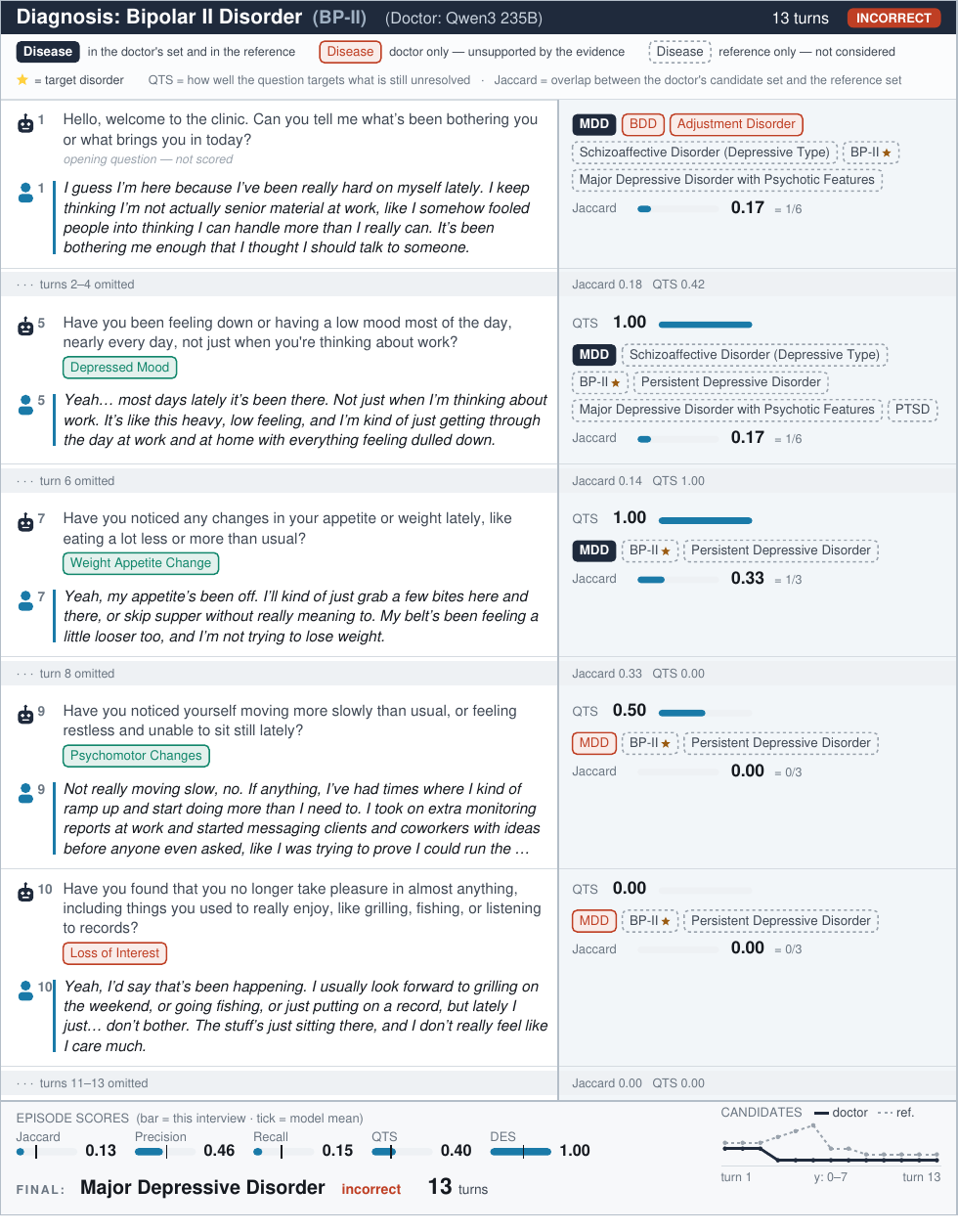}
    \caption{Case study of a diagnostic interview conducted by Qwen3 235B with a patient with bipolar II disorder (BP-II).}
    \label{fig:doctor_case_qwen}
\end{figure*}

\paragraph{Case 3: Incorrect but supported diagnosis.}
\autoref{fig:doctor_case_qwen} illustrates the incorrect-but-supported failure mode discussed in \autoref{sec:correct_not_supported}. The target disorder is bipolar II disorder, but Qwen3 235B narrows its candidate set to major depressive disorder (MDD) alone by Turn 7. At Turn 9, the patient describes taking on extra work and messaging others with ideas unprompted, suggesting hypomanic features, and the reference set accordingly excludes MDD. The doctor nevertheless does not revise its candidate set and continues to ask questions confirming MDD. Its final diagnosis of MDD is incorrect, yet its DES is $1.00$: the doctor acquires sufficient evidence for its own working diagnosis but fails to account for contradicting evidence, consistent with the tendency of Qwen3 235B toward early diagnostic commitment.

\newpage
\raggedbottom

\begin{figure*}[t!]
    \centering
    \includegraphics[width=0.49\linewidth]{Images/Results/v4_radar_gpt-5.6-terra.pdf}
    \includegraphics[width=0.49\linewidth]{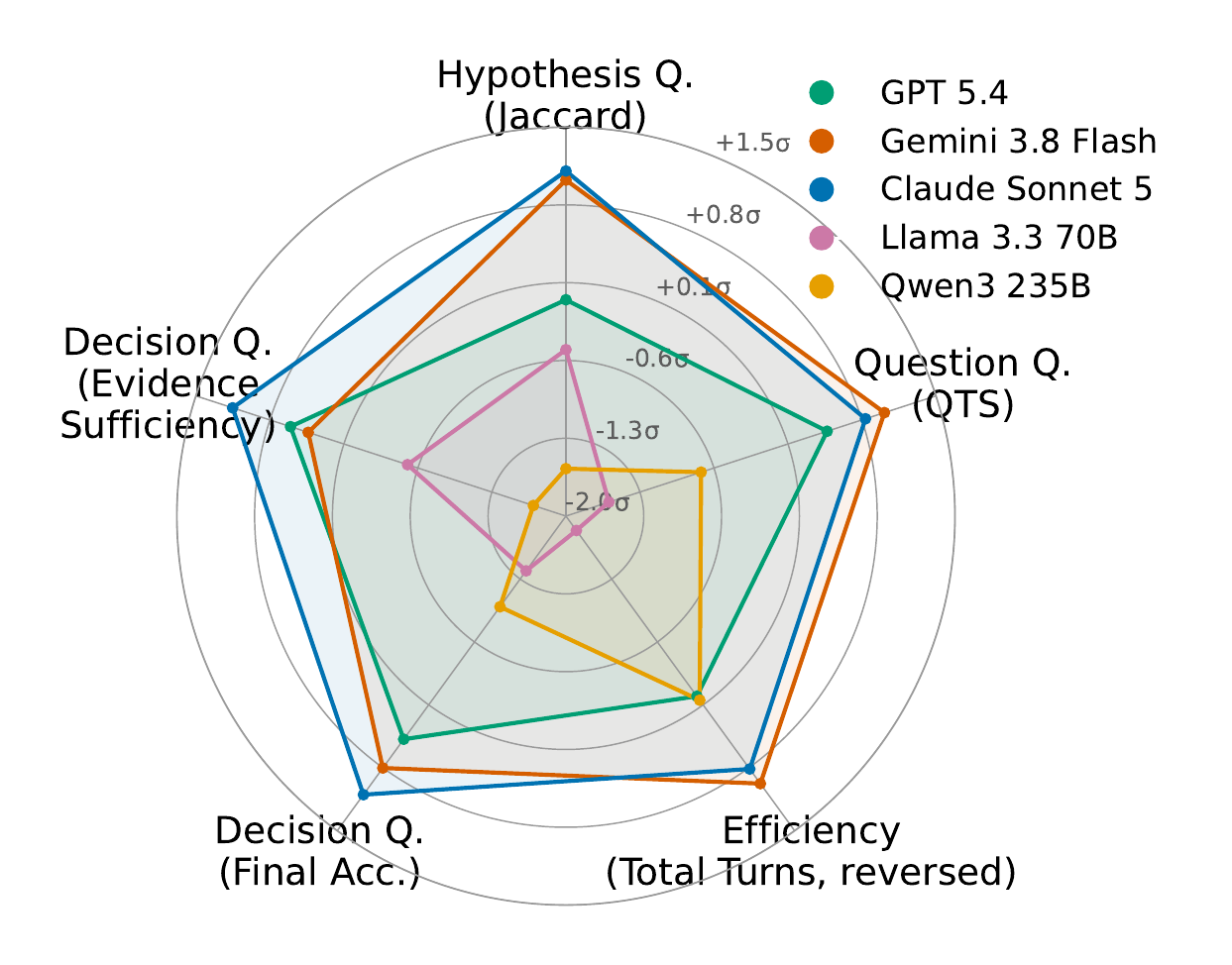}
    \caption{Diagnostic behavior of doctor models across five metrics under the GPT 5.6 Terra (left) and Gemini 3.1 Pro (right) settings. Scores are standardized across models within each setting, and Efficiency is inverted so that higher values indicate fewer turns. The profiles are consistent across the two settings: Claude Sonnet 5 and Gemini 3.8 Flash lead on hypothesis quality, question quality, and efficiency, GPT 5.4 leads on evidence sufficiency, and Llama 3.3 70B and Qwen3 235B remain lower on most dimensions.}
    \vspace{-0.2in}
    \label{fig:overall_profile_append}
\end{figure*}

\section{Experiments and Analysis}

\textcolor{red}{여기 부분. Table 3 서술은 Gemini Pro 실험 결과에 대해서만 서술. 분석은 둘 같이 보여주고 비교. 숫자 바꾸기.}

\subsection{Detailed Results}
\label{app:detailed_results}

\begin{figure*}[!t]
    \centering
    \includegraphics[width=0.49\linewidth]{Images/Results/narrowing_progress_predicted_gpt-5.6-terra.pdf}
    \includegraphics[width=0.49\linewidth]{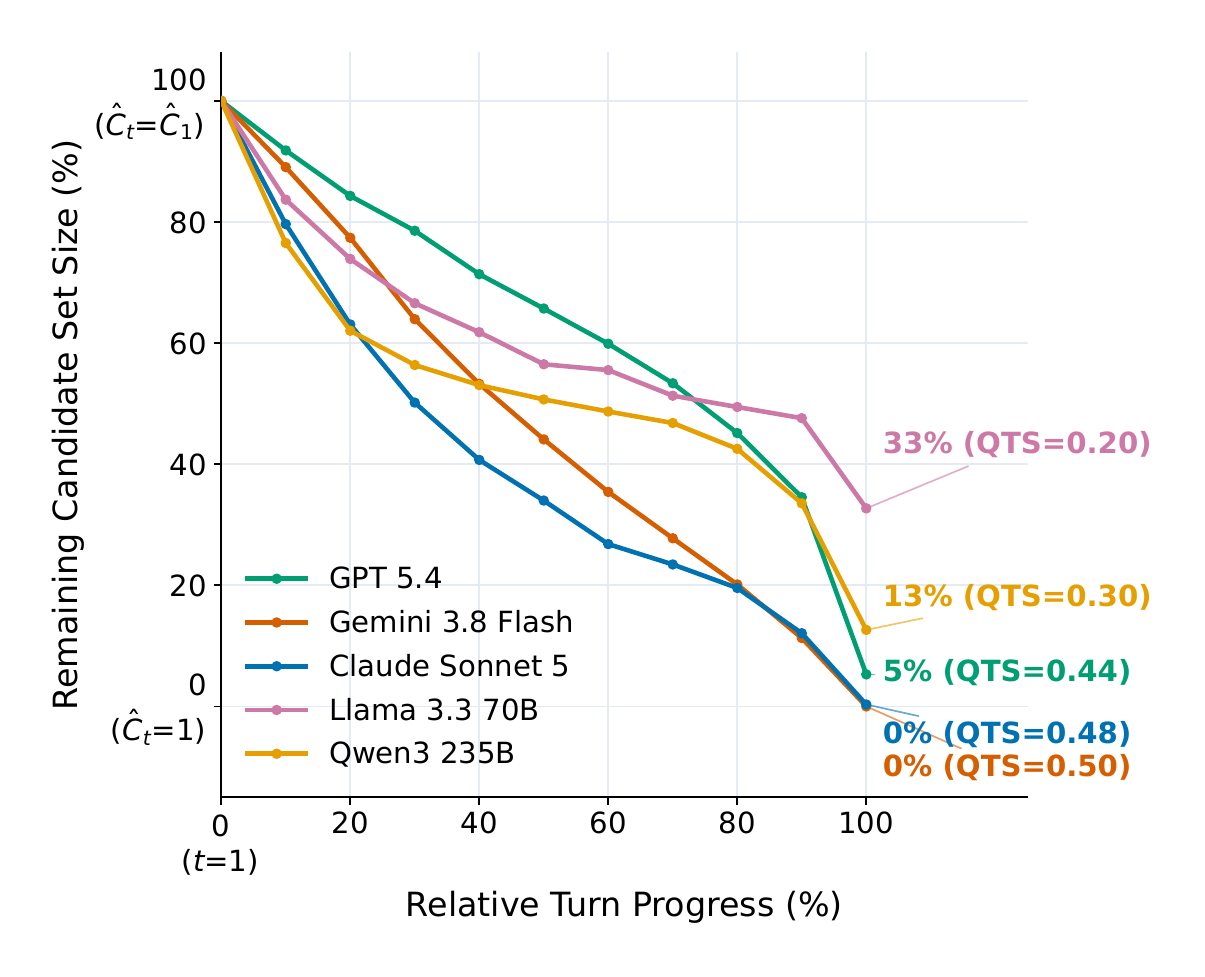}
    \caption{Narrowing progress over relative interview turns and Question Targeting Score (QTS) under the GPT 5.6 Terra (left) and Gemini 3.1 Pro (right) settings. The y-axis shows the remaining reduction to a single candidate, normalized to its initial value (0\% indicates at most one candidate; 100\% indicates the start), along a standardized timeline on the x-axis. Higher-QTS models generally show greater candidate reduction as the interview progresses.}

    \label{fig:narrowing progress_qts_append}
\end{figure*}
\begin{figure*}[t]
    \centering
    \includegraphics[width=0.49\linewidth]{Images//Results/quadrant_flower_gpt-5.6-terra.pdf}
    \includegraphics[width=0.49\linewidth]{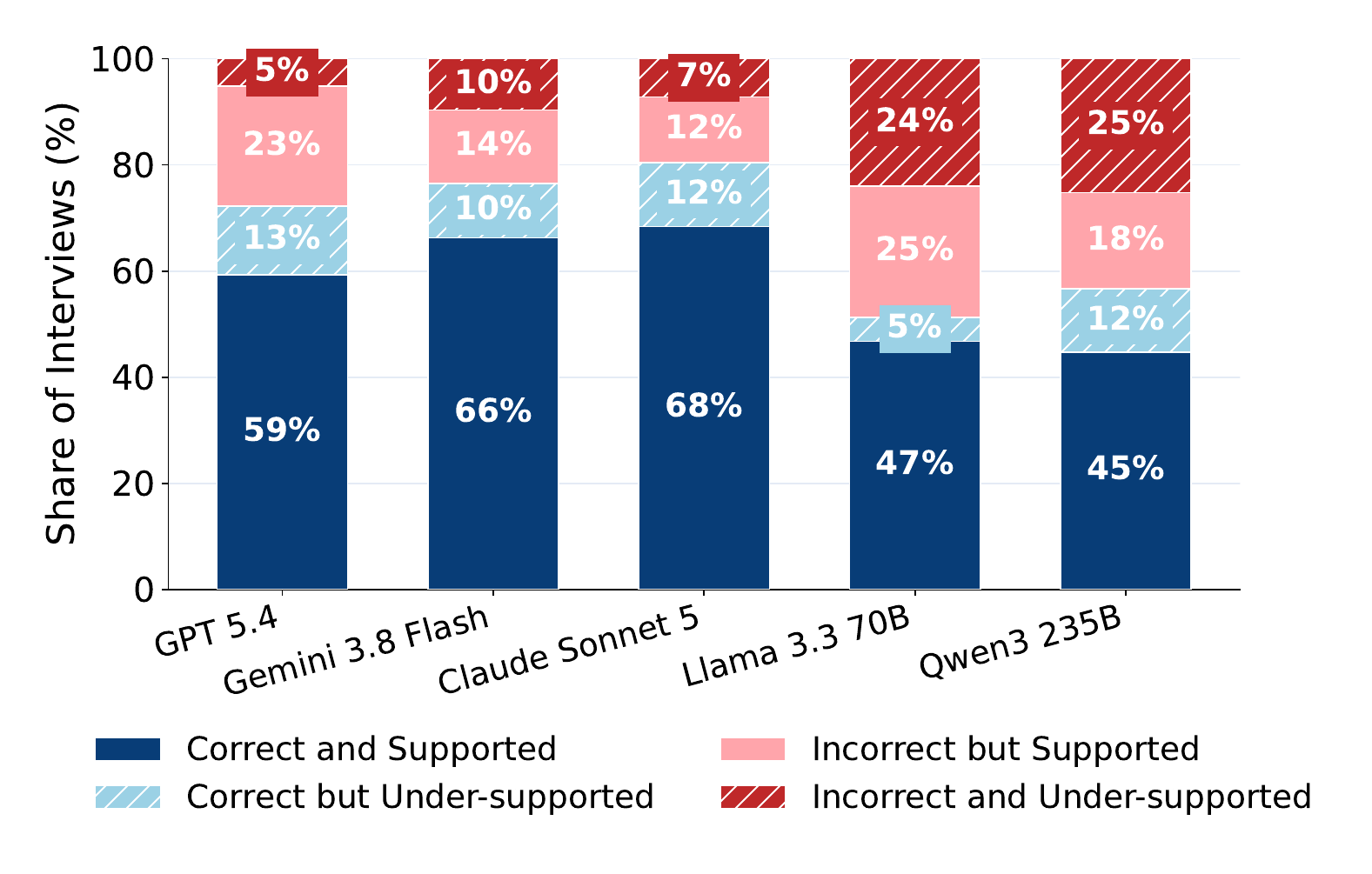}
    \caption{Final diagnostic accuracy and evidence sufficiency across doctor models under the GPT 5.6 Terra (left) and Gemini 3.1 Pro (right) settings. Interviews are grouped by whether the final diagnosis is correct and whether it is sufficiently supported by the acquired evidence; percentages indicate the share of interviews in each category. In both settings, most interviews of the closed-source models are correct and supported ($60$--$68\%$), whereas Llama 3.3 70B and Qwen3 235B produce more incorrect diagnoses, a substantial share of which are nonetheless supported by the acquired evidence ($19$--$30\%$).
    Interviews whose final diagnosis does not map to a MentalKG disorder (and thus lack DES) are excluded.
    }
    \vspace{-0.2in}
    \label{fig:accuracy_vs_reasoning_append}
\end{figure*}

Section~\ref{sec: Experiments} reports results under the GPT 5.6 Terra evaluation setting due to page limit. In this section, we present additional results under the Gemini 3.1 Pro setting and examine whether the main findings are consistent across both settings. In each setting, the same LLM serves as both the patient simulator and evaluation judge. Although the clinical profiles are identical, the two settings may produce different patient responses and interview trajectories. We therefore compare aggregate results and behavioral patterns rather than individual interviews. 



\paragraph{Overall Diagnostic Performance.}
As shown in \autoref{tab:main_table}, under the Gemini 3.1 Pro setting, Claude Sonnet 5 achieves the highest Jaccard (0.49), precision (0.69), DES (0.72), and final accuracy ($77.39\%$). 
Gemini 3.8 Flash achieves the highest QTS (0.50) and conducts the shortest interviews (6.19 turns), while also attaining the highest recall (0.64). 
GPT 5.4 achieves a relatively high DES (0.70) but conducts longer interviews (10.57 turns).
Among the open-source models, Llama 3.3 70B exhibits the lowest QTS (0.20) and the longest interviews (18.89 turns), while Qwen3 235B shows the lowest Jaccard (0.34) and DES (0.60). These multidimensional differences are illustrated in \autoref{fig:overall_profile_append}.

These results are largely consistent with those under the GPT 5.6 Terra setting. Across the five doctor models, Jaccard, QTS, and total turns show identical rankings between the two settings (Spearman $\rho=1.0$), while precision and final accuracy show strong rank agreement ($\rho=0.9$ for both), followed by recall ($\rho=0.8$). 

Nevertheless, minor differences exist between the two settings. All doctor models conduct shorter interviews under Gemini 3.1 Pro, with the largest reduction observed for GPT 5.4 (from 15.06 to 10.57 turns). The highest recall shifts from GPT 5.4 to Gemini 3.8 Flash, while the highest DES shifts from GPT 5.4 to Claude Sonnet 5. However, these ranking changes reflect small differences in metric values rather than substantial shifts in model performance. Overall, the relative strengths and diagnostic behavior patterns of the doctor models remain largely consistent across the two evaluation settings.

\paragraph{\dqq{Question Quality and Candidate Narrowing.}}

\autoref{fig:narrowing progress_qts_append} compares how doctor-reported candidate sets narrow across interview progress under both evaluation settings. The y-axis represents the normalized fraction of the reduction from the initial candidate set to a single candidate that remains ($\textstyle \frac{\hat{C}_1 - \hat{C}_t -1}{\hat{C}_1-1}$). Interviews beginning with at most one candidate ($\textstyle \hat{C}_1 = 1$) are excluded, and trajectories are interpolated over relative interview progress before averaging. Empty candidate sets are clipped to $0\%$, so a low remaining fraction does not necessarily indicate convergence to the correct diagnosis.


Under Gemini 3.1 Pro, Claude Sonnet 5 and Gemini 3.8 Flash reach low remaining fractions by the end of the interview, whereas Llama 3.3 70B retains the highest fraction. This pattern is consistent with the GPT 5.6 Terra setting. 

\begin{figure*}[th!]
    \centering
    \includegraphics[width=0.49\linewidth]{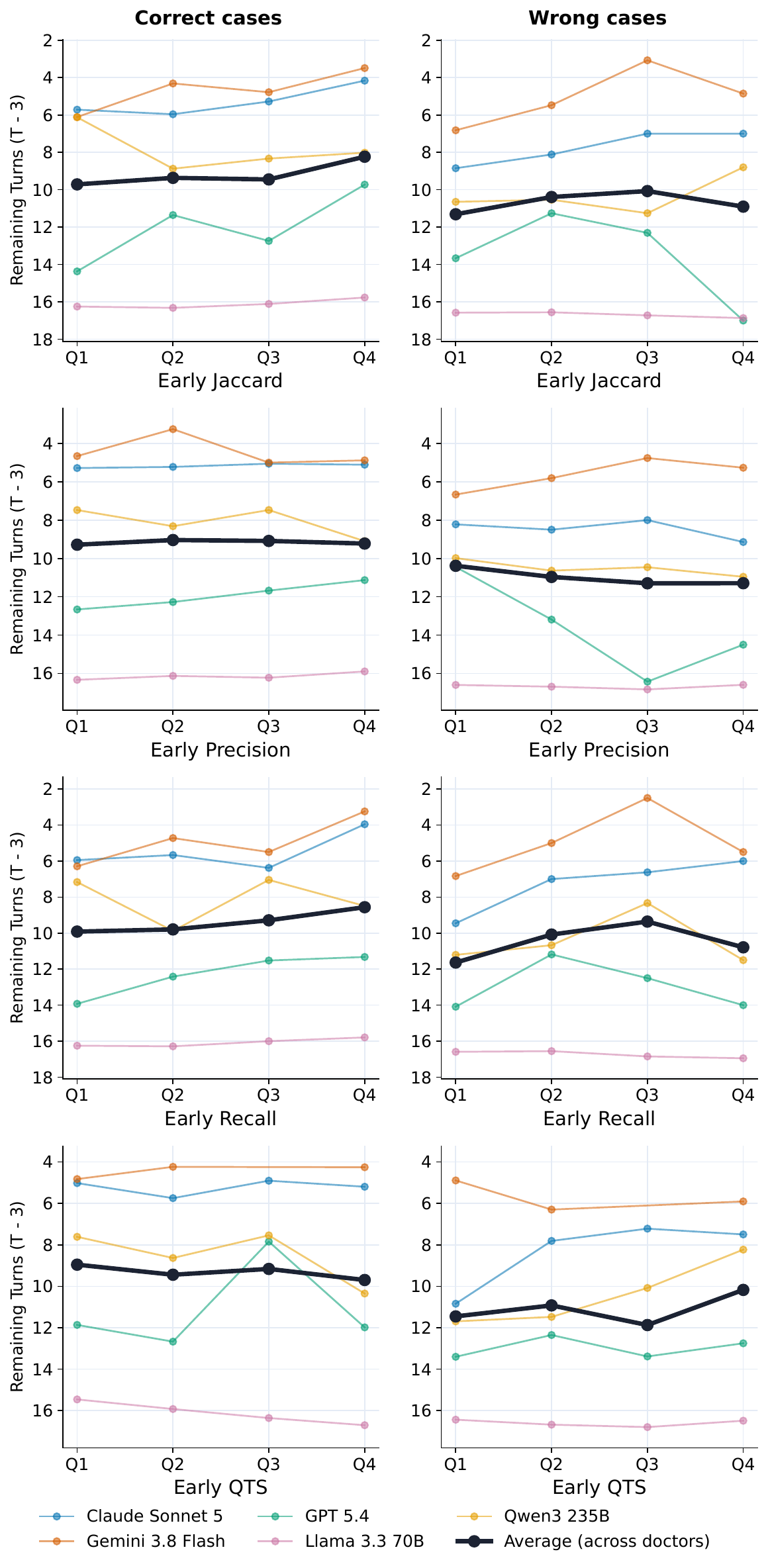}
    \hfill
    \includegraphics[width=0.49\linewidth]{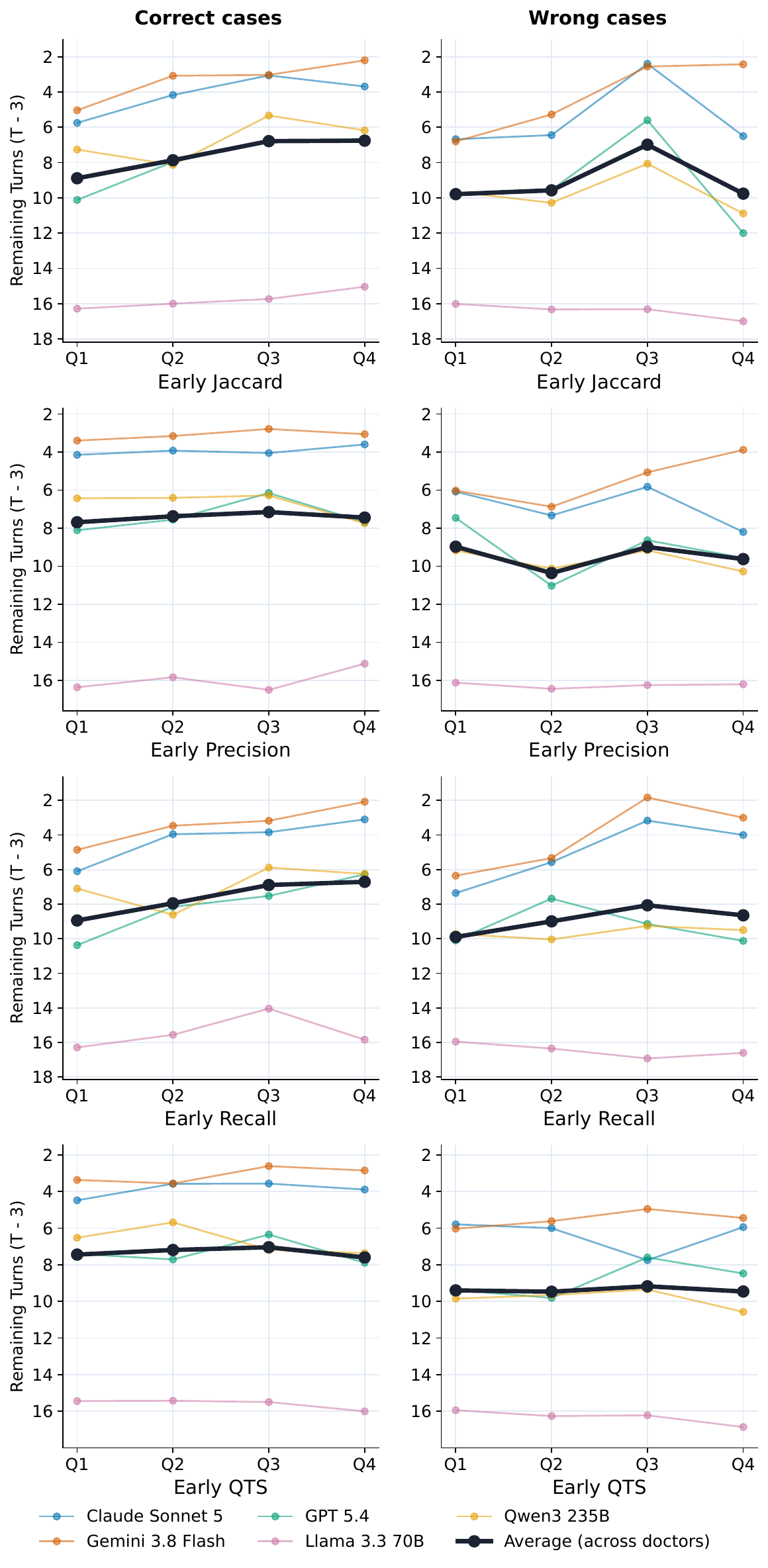}
    \caption{Early diagnostic quality and downstream interview efficiency under the GPT 5.6 Terra (left two columns) and Gemini 3.1 Pro (right two columns) settings. Within each setting, interviews are grouped into quartiles (Q1: lowest, Q4: highest) by each metric averaged over the first three turns, separately for correct and incorrect diagnoses. Lines show the mean number of turns remaining after Turn 3 ($T-3$) for each doctor model, and the thick black line shows the average across doctors; the y-axis is inverted so that higher points indicate fewer remaining turns. In both settings, higher Early Jaccard and Recall are associated with shorter interviews in correct cases, while Early Precision and QTS show no clear trend, consistent with \autoref{fig:efficiency}.}
    \vspace{-0.2in}
    \label{fig:dhq_efficiency_append}
\end{figure*}

\paragraph{\de{Early Hypothesis Quality and Downstream Efficiency.}}
\autoref{fig:dhq_efficiency_append} shows, for interviews longer than three turns, the relationship between each metric averaged over the first three turns and the number of remaining turns ($T-3$). Quartile boundaries are defined within each setting and doctor model by pooling correct and incorrect cases, and the same boundaries are used to display the two outcome groups separately. The thick line is the equal-weight average of the per-model means in each quartile.


In the correct cases of both settings, higher quartiles of Early Jaccard and Early Recall are generally associated with fewer remaining turns on average across models. In contrast, Early Precision and Early QTS show no consistent trend in the same direction, and the incorrect cases show no monotonic relationship across quartiles. This mirrors the main-text finding (\autoref{fig:efficiency}) that, among successful interviews, the quality of the early candidate set---particularly its coverage of plausible candidates---is associated with shorter subsequent interviews. This aggregate pattern, however, does not hold uniformly for every doctor model and does not imply a causal effect.


\paragraph{\ddq{Final Accuracy and Evidence Sufficiency.}}
\autoref{tab:main_table} shows that final accuracy and DES capture different aspects of diagnostic decisions. Claude Sonnet 5 achieves the highest accuracy in both settings, whereas GPT 5.4 achieves the highest DES. Llama 3.3 70B has higher DES than Qwen3 235B ($0.62$ vs.\ $0.54$; $0.65$ vs.\ $0.59$) yet lower accuracy ($41.45\%$ vs.\ $46.09\%$; $45.80\%$ vs.\ $50.87\%$). In other words, how well the acquired evidence satisfies the requirements of the predicted disorder is distinct from whether the doctor reaches the correct disorder.

\autoref{fig:accuracy_vs_reasoning_append} jointly categorizes interviews by these two aspects. Here, ``Supported'' denotes a DES of at least $0.5$; it is an analytic threshold and does not mean that all diagnostic requirements are satisfied. Both correct but under-supported and incorrect but supported cases appear in both settings.

DES is undefined when the doctor's final diagnosis code does not map to any of the 23 MentalKG disorders (e.g., reporting the symptom code R63.0 instead of a disorder code). Such interviews are counted as incorrect in final accuracy, but because there is no disorder whose requirements can be assessed, they are excluded from the DES averages in \autoref{tab:main_table} and from \autoref{fig:accuracy_vs_reasoning} and \autoref{fig:accuracy_vs_reasoning_append}. Excluded interviews account for $3.2$--$12.5\%$ of each doctor model's interviews ($11$--$43$ interviews), most frequently for Llama 3.3 70B and Qwen3 235B. Since all excluded interviews are incorrect, the combined share of the correct categories in the figures is $2.5$--$5.9$ percentage points higher than the final accuracy in the table, and the category shares should therefore not be equated directly with the reported accuracy.

\paragraph{Conclusions That Hold Across Evaluation Settings.}
Across the two evaluation settings, the main relative characteristics of each doctor model persist: Claude's high hypothesis precision and final accuracy, Gemini's high QTS and short interviews, GPT's high DES, Llama's low QTS and long interviews, and Qwen's low Jaccard and DES. At the same time, absolute performance and some rankings vary. These results therefore support that the main behavioral patterns are reproduced under the two patient--judge combinations we evaluate. We do not extend this to independence from the patient or judge model individually, or to invariance across evaluation settings in general.

\end{document}